\batchmode
\makeatletter
\def\input@path{{"C:/Trabajo laptop/Mis articulos/Drone traffic monitoring/"}}
\makeatother
\documentclass[british,english]{IEEEtran}
\usepackage[T1]{fontenc}
\usepackage[latin9]{inputenc}
\usepackage{array}
\usepackage{amsmath}
\usepackage{amsthm}
\usepackage{amssymb}
\usepackage{graphicx}
\PassOptionsToPackage{normalem}{ulem}
\usepackage{ulem}

\makeatletter

\providecommand{\tabularnewline}{\\}

\theoremstyle{plain}
\newtheorem{thm}{\protect\theoremname}
\theoremstyle{plain}
\newtheorem{lem}[thm]{\protect\lemmaname}

\usepackage{cite} 
\usepackage[margin=8pt,font=footnotesize]{caption}
\usepackage{algorithm}
\usepackage{algpseudocode}
\usepackage{amsmath}  
\allowdisplaybreaks 

\makeatother

\usepackage{babel}
\addto\captionsbritish{\renewcommand{\lemmaname}{Lemma}}
\addto\captionsbritish{\renewcommand{\theoremname}{Theorem}}
\addto\captionsenglish{\renewcommand{\lemmaname}{Lemma}}
\addto\captionsenglish{\renewcommand{\theoremname}{Theorem}}
\providecommand{\lemmaname}{Lemma}
\providecommand{\theoremname}{Theorem}

\begin{document}
\title{Trajectory Poisson multi-Bernoulli mixture filter for traffic monitoring
using a drone}
\author{Ángel F. García-Fernández and Jimin Xiao\thanks{A. F. García-Fernández is with the Department of Electrical Engineering and Electronics, University of Liverpool, Liverpool L69 3GJ, United Kingdom, and also with the ARIES Research Centre, Universidad Antonio de Nebrija,  Madrid, Spain (email: angel.garcia-fernandez@liverpool.ac.uk). J. Xiao is with the Department of Electrical and Electronic Engineering, Xi'an Jiaotong-Liverpool University, Suzhou 215123, China (email: jimin.xiao@xjtlu.edu.cn). The authors would like to thank support by the Royal Society Research Grant RGS-R1-221049.}}
\maketitle
\begin{abstract}
This paper proposes a multi-object tracking (MOT) algorithm for traffic
monitoring using a drone equipped with optical and thermal cameras.
Object detections on the images are obtained using a neural network
for each type of camera. The cameras are modelled as direction-of-arrival
(DOA) sensors. Each DOA detection follows a von-Mises Fisher distribution,
whose mean direction is obtain by projecting a vehicle position on
the ground to the camera. We then use the trajectory Poisson multi-Bernoulli
mixture filter (TPMBM), which is a Bayesian MOT algorithm, to optimally
estimate the set of vehicle trajectories. We have also developed a
parameter estimation algorithm for the measurement model. We have
tested the accuracy of the resulting TPMBM filter in synthetic and
experimental data sets. 
\end{abstract}


\section{Introduction}

Traffic monitoring consists of obtaining relevant information on the
traffic in a given area based on sensor data \cite{Zhou07,Du15}.
It provides key information that enables the improvement of the roadway
system, and  traffic law enforcement \cite{Won20,Pan21}. There are
different types of sensors used for traffic monitoring. For example,
there are sensors that are located under or on the road: piezoelectric
sensors, magnetometers, vibration sensors and loop detectors \cite{Won20}.
These sensors can provide valuable information on the number of vehicles
and their types. However, they are difficult to install and only provide
information at a given location. Other types of sensors for traffic
monitoring are cameras, radars and lidars, which can be mounted next
to the road or in another vehicle \cite{Won20}. 

This paper focuses on traffic monitoring using a drone, also called
an unmanned aerial vehicle \cite{Vahidi19,Barmpounakis19,Huang21}.
A drone is convenient device to gather traffic data as it can be easily
deployed and equipped with different sensors, such as cameras and
lidars. We proceed to review the literature on traffic monitoring
using drones equipped with cameras. 

Traffic monitoring using a drone with cameras requires an object detection
algorithm to detect vehicles on the roads and extract meaningful analytics
\cite{Ke19}. It is also possible to improve object detection by adding
road segmentation and only detect vehicles on the roads \cite{Kyrkou18,Makrigiorgis22}.
In \cite{Du18}, several vehicle trackers on the image plane are evaluated.
In \cite{Makrigiorgis22}, vehicle tracking is performed on the image
plane using Kalman filters and data association methods. Velocity
on the ground plane is obtained projecting the pixels to the ground
plane using the ground sample distance. In \cite{Ke19}, traffic flow
estimation is performed by projecting the downward-facing image and
applying the \foreignlanguage{british}{Kanade--Lucas--Tomasi (KLT)}
algorithm \cite{Lucas81,Maggio_book11} to estimate the background
motion in low traffic. Marked geo-referenced points can also be used
to accurately project pixels to the ground plane and obtain traffic
statistics \cite{Barmpounakis19}. In \cite{Wang16}, vehicles are
detected on the image, and the optical flow using the KLT is obtained
to perform the tracking. Road features are used to calculate the camera
motion and pixel projection to the ground plane is achieved through
an homography transformation obtained with matching points. 

In this paper, we propose to use Bayesian multi-object tracking (MOT)
algorithms to perform traffic monitoring directly on the ground plane.
In Bayesian MOT\footnote{An online course on Bayesian MOT is https://www.youtube.com/channel/UCa2-fpj6AV8T6JK1uTRuFpw. },
we require a multi-object dynamic model, which includes object births,
single-object dynamics and deaths, and a multi-object measurement
model \cite{Mahler_book14}. We consider the standard point-object
dynamic and detection-based measurement models, such that object birth
follows a Poisson point process (PPP), and each object may generate
at most one detection. In this framework, all information of interest
about the object trajectories is included in the density of the set
of (vehicle) trajectories given past and current measurements \cite{Angel20_b}.
This posterior density can then be used to obtain the required traffic
analytics. 

The posterior density is a Poisson multi-Bernoulli mixture (PMBM)
density \cite{Williams15b,Granstrom18,Li21}. The PMBM density is
the union of an independent PPP and a multi-Bernoulli mixture (MBM).
The PPP contains information on potential trajectories that have not
yet been detected, and is useful for example in search-and-track sensor
management \cite{Bostrom-Rost21}. The MBM contains the information
on potential trajectories that have been detected at some point, as
well as their data associations (global hypotheses). This PMBM density
is computed recursively via the prediction and update steps of the
trajectory PMBM (TPMBM) filter \cite{Granstrom18,Angel20_e}.  

Object detections from the images are obtained using a deep learning
detector \cite{Scheidegger18}. Then, to apply the TPMBM filter to
these detections, we require a measurement model. There are several
models available in the literature. In \cite{Scheidegger18}, the
camera is modelled as a range-bearings sensor with additive Gaussian
noise, with the neural network also providing depth information. In
\cite{Helgesen22}, the detection is converted to an azimuth measurement,
which is assumed to be Gaussian distributed. In \cite{Helgesen20},
the camera detection is mapped to a position with zero elevation.
The measurement model is then a positional measurement on the ground
plane with zero-mean Gaussian noise whose covariance matrix is calculated
through the Jacobian of the projection mapping. It is also possible
to use the unscented transform \cite{Julier04} to determine this
covariance matrix \cite{Helgesen23}. The same type of approach is
followed in \cite{Sharma18} but with predefined diagonal covariance
matrix for the noise.

In this paper, we model the camera as a direction-of-arrival (DOA)
sensor, such that each object detection is a DOA. We use a von-Mises
Fisher (VMF) distribution for DOAs, as it is a mathematically principled
distribution for directional data and has benefits over Gaussian distributions
for this type of data \cite{Mardia_book00,Angel19_c}. For example,
VMF distributions intrinsically deal with DOAs and do not require
angle wrapping or external approaches to design filtering algorithms
\cite{Crouse15c}. Then, using geometrical considerations, we obtain
the distribution of the DOA measurement given a vehicle state, and
also the distribution of the clutter measurements. 

A challenge to apply the above camera-detection modelling to TPMBM
filtering is to deal with the non-linear/non-Gaussian measurement
updates for trajectories. One possibility is to use particle filtering
techniques \cite{Arulampalam02}. A computationally cheaper alternative
is to use a non-linear Kalman filter, such as the extended Kalman
filter (EKF) \cite{Sarkka_book13} or the sigma-point Kalman filters
such as the unscented Kalman filter \cite{Julier04} and the cubature
Kalman filter \cite{Sarkka_book13}. These filters implicitly perform
a linearisation of the measurement function to obtain a Gaussian approximation
to the posterior. The sigma-point Kalman filters have the following
benefits over the EKF: they do not require the calculation of the
Jacobian of the measurement function, and they perform the linearisation
taking into account the uncertainty in the predicted density. The
EKF and sigma-point Kalman filters work well for mild measurement
nonlinearities \cite{Morelande13}. However, they do not take into
account the information provided by the measurement to perform the
linearisation, which implies that there is room for improvement with
stronger nonlinearities. 

To improve performance, we therefore use the iterated posterior linearisation
filter (IPLF) to perform the updates for the single-trajectory densities.
The IPLF aims to obtain the best linearisation of the measurement
function, as well as the resulting linearisation error, taking into
account the observed measurement \cite{Angel15_c}. This property
enables the IPLF to outperform the EKF and UKF if the update is sufficiently
non-linear. The IPLF consists of an iterated procedure in which we
linearise the measurement function in the area where the current approximation
of the posterior lies. The IPLF can be implemented via sigma-points
and its first iteration corresponds to a sigma-point Kalman filter
update. In particular, we use the IPLF for VMF-distributed measurements
and single-trajectory densities \cite{Angel22_c}.  

The contributions of this paper are:
\begin{itemize}
\item A camera measurement model, suitable for traffic monitoring with a
drone, based on a deep learning detector and the VMF distribution. 
\item The TPMBM filter implementation with the VMF camera measurement model,
the IPLF to perform the single-trajectory updates, and a suitable
birth model.
\item A measurement model parameter estimation algorithm based on optimising
the likelihood, augmented with auxiliary variables representing data
association hypotheses. The algorithm iteratively optimises the auxiliary
variables, the clutter rate, probability of detection and the concentration
parameter of the VMF distribution. 
\item The evaluation of the TPMBM filter via synthetic and experimental
data using optical and thermal cameras on a Parrot Anafi USA drone
\cite{Parrot2022}.
\end{itemize}
The rest of the paper is organised as follows. The problem formulation
is provided in Section \ref{sec:System-description-and}. In Section
\ref{sec:Camera-as-direction-of-arrival}, we present the multi-object
measurement model. The measurement model parameter estimation algorithm
is proposed in Section \ref{sec:Measurement-parameter-estimation}.
The multi-object dynamic model is introduced in Section \ref{sec:Multi-object-dynamic-model}.
The main aspects of the Gaussian TPMBM implementation are addressed
in Section \ref{sec:Gaussian-TPMBM-implementation}. Finally, experimental
results and conclusions are presented in Sections \ref{sec:Experiments}
and \ref{sec:Conclusions}, respectively.

\section{System description, problem formulation, and TPMBM filtering\label{sec:System-description-and}}

In this section we first describe the drone specifications in Section
\ref{subsec:Drone-specifications}. Then, we present the considered
coordinate systems in Section \ref{subsec:Coordinate-systems}. We
formulate the MOT problem from a Bayesian perspective in Section \ref{subsec:Problem-formulation}
and review the TPMBM filter in Section \ref{subsec:TPMBM-filtering}. 

\subsection{Drone specifications\label{subsec:Drone-specifications}}

In our experiments, we use a Parrot Anafi USA drone \cite{Parrot2022}.
While the proposed algorithm is of general applicability, it is important
to understand from the beginning what type of information is provided
by the drone. The drone is equipped with several sensors for accurate
positioning: satellite navigation (GPS, GLONASS and GALILEO), barometer,
magnetometer, vertical camera and ultra-sonar, 2 inertial measurement
units, 2 accelerometers and 2 gyroscopes. The drone is also equipped
with two optical cameras and one thermal camera. We record video in
either optical or thermal camera mode. 

For each video frame, we can obtain the drone elevation, GPS location,
attitude of the camera (in the form of a quaternion), and the FoV
in degrees. The FoV depends on the user-defined zoom, which we assume
fixed in time for simplicity. The FoV, frames per second (fps) and
image size in pixels used for the optical and thermal camera modes
are shown in Table \ref{tab:Video-specifications}. 

\begin{table}
\caption{\label{tab:Video-specifications}Video specifications}

\centering{}%
\begin{tabular}{c|cc}
\hline 
Parameter &
Optical &
Thermal\tabularnewline
\hline 
FoV &
$\left[69^{\circ},42.27^{\circ}\right]$ &
$\left[46.14^{\circ},36.75^{\circ}\right]$\tabularnewline
Fps &
30 &
8.6\tabularnewline
Image size &
$1920\times1080$ &
$1190\times928$\tabularnewline
\hline 
\end{tabular}
\end{table}

\subsection{Coordinate systems\label{subsec:Coordinate-systems}}

In this work, we use three coordinates systems: the world geodetic
system (WGS) 84 format (which is the standard reference system for
GPS navigation) \cite{Cai_book11}, a local coordinate system for
each flight, and a camera coordinate system. We proceed to explain
these.

The drone coordinates in the WGS84 format, i.e., its latitude, longitude
and altitude, which are obtained via satellite navigation, are recorded
for each frame. For each flight, we use a local Cartesian coordinate
system with the center located on the ground with the initial latitude
and longitude of the drone. In this coordinate system, the $x$-axis
points to the East, the $y$-axis points to the North and the $z$-axis
points down. This corresponds to the initial drone north-east-down
(NED) coordinate system located on the ground \cite{Cai_book11}.

\begin{figure}
\begin{centering}
\includegraphics[scale=0.5]{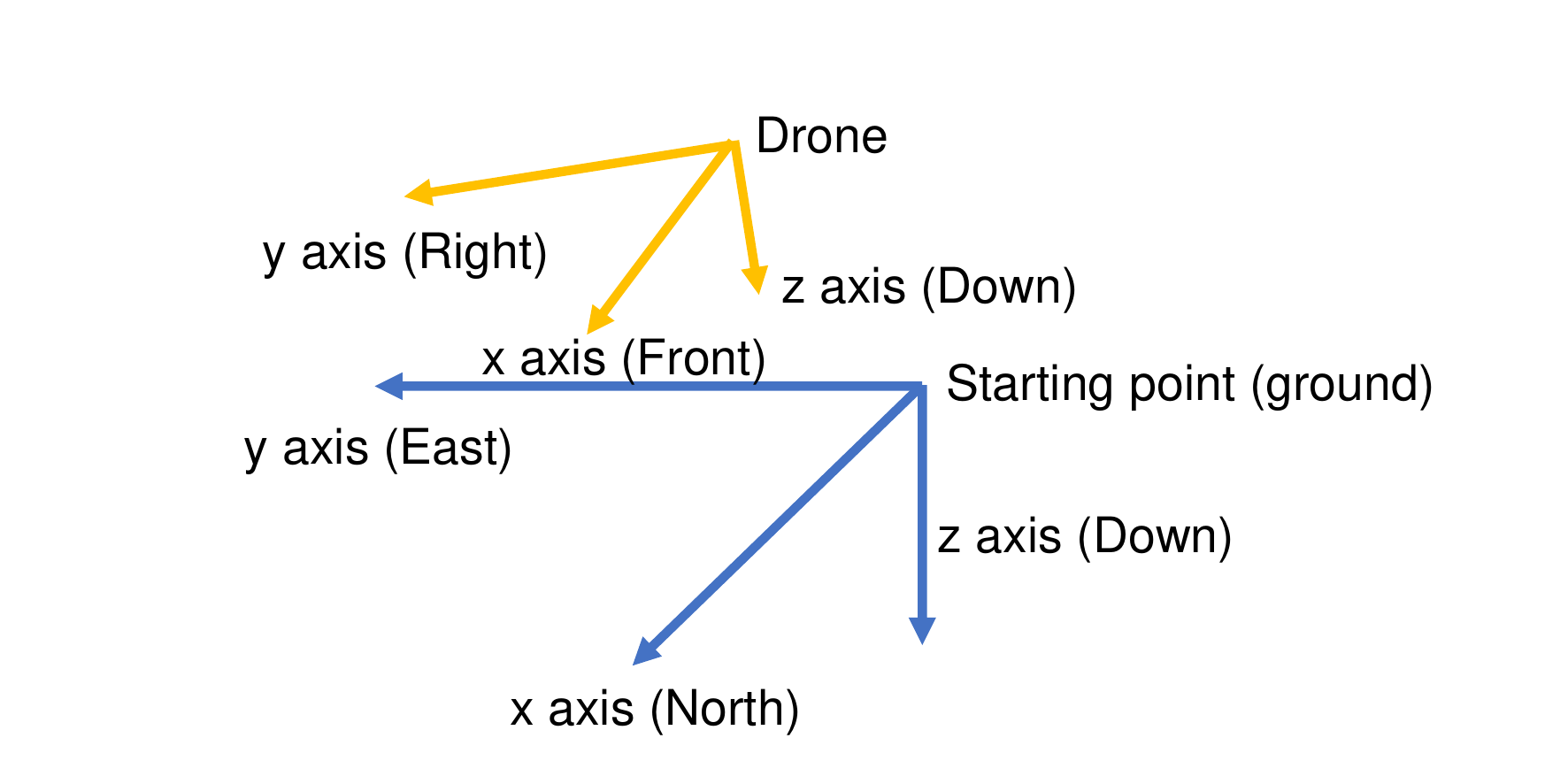}
\par\end{centering}
\caption{\label{fig:Local-coordinate-systems}Cartesian coordinate systems.
Local NED coordinate system with origin located on the ground and
the initial latitude and longitude of the drone (blue). Camera coordinate
system on the drone (yellow). The vehicles are tracked in the local
coordinate system. }

\end{figure}

We also use a Cartesian coordinate system on the camera. In this coordinate
system, the $x$-axis points ahead of the camera, the $y$-axis points
to the right of the $x$-axis (in the direction of view), and the
$z$-axis down, see Figure \ref{fig:Local-coordinate-systems}. The
last two coordinate systems are used because the drone directly provides
us with the rotation of this camera reference frame w.r.t. the local
NED coordinate system.

\subsection{Problem formulation\label{subsec:Problem-formulation}}

We model the MOT problem using a Bayesian approach \cite{Mahler_book14}.
Given the Bayesian modelling, our objective is to compute the posterior
density of the set of all the vehicle trajectories (in the local coordinate
system) that have been in the FoV \cite{Angel20_b}. We proceed to
explain the models and the variables in the problem formulation

We make the assumption that all the objects (vehicles) are at a constant
elevation in the local coordinate system. An object state in the local
coordinate system is denoted by $x=\left[p_{x},v_{x},p_{y},v_{y}\right]^{T}\in\mathbb{R}^{n_{x}}$,
with $n_{x}=4$, and contains its position and velocity in the local
coordinate system. The set of objects at the current time step $k$
is $\mathbf{x}_{k}=\left\{ x_{k}^{1},...,x_{k}^{n_{k}}\right\} $,
where $n_{k}$ is the number of objects. 

At each time step, each object $x\in\mathbf{x}_{k}$ can survive to
the next time step with probability $p^{S}$ of survival and move
with a Markovian transition density $g\left(\cdot\left|x\right.\right)$.
In addition, independent new objects may be born following a PPP with
intensity $\lambda_{k}^{B}\left(\cdot\right)$. According to this
model, a vehicle trajectory can be represented as a variable $X=\left(t,x^{1:\nu}\right)$,
where $t$ is the birth time step, $\nu$ is the trajectory length
and $x^{1:\nu}=\left(x^{1},...,x^{\nu}\right)$ denotes the sequence
of length $\nu$ with the object states at consecutive time steps
\cite{Angel20_b}. The set of all trajectories ever present in the
FoV up to the current time step $k$ is then $\mathbf{X}_{k}=\left\{ X_{k}^{1},...,X_{k}^{n_{k}^{t}}\right\} $,
where $n_{k}^{t}$ is the number of trajectories.

At each time step, each object $x\in\mathbf{x}_{k}$ may be detected
with a probability $p^{D}$ and generate a measurement $z\in\mathbb{R}^{n_{z}}$
with density $l\left(\cdot|\mathbf{x}_{k}\right)$. The set of measurements
$\mathbf{z}_{k}$ includes object-detected measurements and clutter
measurements, which are distributed following a PPP with intensity
$\lambda^{C}\left(\cdot\right)$.

\subsection{TPMBM filtering\label{subsec:TPMBM-filtering}}

Using the Bayesian prediction and update steps \cite{Mahler_book14},
we can calculate the posterior and predicted densities of $\mathbf{X}_{k'}$,
denoted by $f_{k'|k}\left(\cdot\right)$ with $k'\in\left\{ k,k+1\right\} $,
given the measurements up to time step $k$. These densities can be
calculated by the TPMBM filtering recursion and are PMBMs of the form
\cite{Granstrom18}
\begin{align}
f_{k'|k}\left(\mathbf{X}_{k'}\right) & =\sum_{\mathbf{Y}\uplus\mathbf{W}=\mathbf{X}_{k'}}f_{k'|k}^{p}\left(\mathbf{Y}\right)f_{k'|k}^{mbm}\left(\mathbf{W}\right),\label{eq:TPMBM_original_mult}\\
f_{k'|k}^{p}\left(\mathbf{X}_{k'}\right) & =e^{-\int\lambda_{k'|k}\left(X\right)dX}\left[\lambda_{k'|k}\left(\cdot\right)\right]^{\mathbf{X}_{k'}},\label{eq:PPP_intensity}\\
f_{k'|k}^{mbm}\left(\mathbf{X}_{k'}\right) & =\sum_{a\in\mathcal{A}_{k'|k}}w_{k'|k}^{a}\sum_{\uplus_{l=1}^{n_{k'|k}}\mathbf{X}^{l}=\mathbf{X}_{k'}}\prod_{i=1}^{n_{k'|k}}f_{k'|k}^{i,a^{i}}\left(\mathbf{X}^{i}\right),\label{eq:MBM_detected}
\end{align}
where (\ref{eq:TPMBM_original_mult}) is the convolution formula for
independent random finite sets, where we sum over all disjoint and
possibly empty sets $\mathbf{Y}$ and $\mathbf{W}$ such that $\mathbf{Y}\cup\mathbf{W}=\mathbf{X}_{k'}$
\cite{Mahler_book14}. That is, (\ref{eq:TPMBM_original_mult}) represents
the union of an independent PPP $f_{k'|k}^{p}\left(\cdot\right)$
with intensity $\lambda_{k'|k}\left(\cdot\right)$, representing trajectories
that remain undetected, and an MBM $f_{k'|k}^{mbm}\left(\cdot\right)$,
representing trajectories that have been detected at some point. The
multi-object exponential, used in (\ref{eq:PPP_intensity}), is defined
as $h^{\mathbf{X}}=\prod_{X\in\mathbf{X}}h\left(X\right)$ where $h$
is a real-valued function and $h^{\emptyset}=1$ by convention.

Equation (\ref{eq:MBM_detected}) sums over all global data association
hypotheses $a\in\mathcal{A}_{k'|k}$, which are data-to-data hypotheses
and have weight $w_{k'|k}^{a}$. Each global hypothesis has an associated
multi-Bernoulli with $n_{k'|k}$ Bernoulli densities. Also, $a=\left(a^{1},...,a^{n_{k'|k}}\right)$,
where $a^{i}\in\left\{ 1,...,h^{i}\right\} $ is the index to the
local hypothesis for the $i$-th Bernoulli and $h^{i}$ is the number
of local hypotheses. The $i$-th Bernoulli with local hypothesis $a^{i}$
has a density 
\begin{align}
f_{k'|k}^{i,a^{i}}\left(\mathbf{X}\right) & =\begin{cases}
1-r_{k'|k}^{i,a^{i}} & \mathbf{X}=\emptyset\\
r_{k'|k}^{i,a^{i}}p_{k'|k}^{i,a^{i}}\left(X\right) & \mathbf{X}=\left\{ X\right\} \\
0 & \mathrm{otherwise},
\end{cases}\label{eq:Bernoulli_density_filter}
\end{align}
where $r_{k'|k}^{i,a^{i}}$ is the probability of existence and $p_{k'|k}^{i,a^{i}}\left(\cdot\right)$
its single-trajectory density, representing the trajectory information
for this local hypothesis. 

A detailed description of (\ref{eq:TPMBM_original_mult})-(\ref{eq:Bernoulli_density_filter})
and the TPMBM filtering recursion can be found in \cite{Granstrom18,Angel20_e}.
To apply TPMBM filtering to drone traffic monitoring, we need to formulate
the task using the models in Section \ref{subsec:Problem-formulation},
which we proceed to do in the following sections.

\section{DOA multi-object measurement model\label{sec:Camera-as-direction-of-arrival}}

In this section, we present the multi-object measurement model characterising
a camera as a DOA sensor. We first review the VMF distribution in
Section \ref{subsec:Von-Mises-Fisher-distribution}. Then, we explain
the relation between a pixel and a DOA in Section \ref{subsec:Pixel-DOA-equivalence}.
We then provide the object-generated measurement model and the clutter
model in Sections \ref{subsec:Object-generated-measurement} and \ref{subsec:Clutter-generated-measurement}.
The overall measurement model is explained in Section \ref{subsec:Multi-object-measurement-model}. 

\subsection{Von Mises-Fisher distribution\label{subsec:Von-Mises-Fisher-distribution}}

The VMF distribution is a convenient and mathematically principled
distribution for DOAs \cite{Mardia_book00}. As we deal with a 3-D
model, a DOA measurement is represented as a unit vector $z$ corresponding
to a point in the sphere $S^{2}=\left\{ z:\:z^{T}z=1,\,z\in\mathbb{R}^{3}\right\} $.
The VMF distribution is unimodal and parameterised by a mean direction
$\mu$ and a concentration parameter $\kappa$. Its density with respect
to the uniform distribution is \cite[Eq. (9.3.4)]{Mardia_book00}
\begin{align}
\mathcal{V}\left(z;\mu,\kappa\right) & =\frac{\left(\frac{\kappa}{2}\right)^{1/2}}{\Gamma\left(3/2\right)\mathrm{I}_{1/2}\left(\kappa\right)}\exp\left(\kappa\mu^{T}z\right)\chi_{\left\Vert z\right\Vert =1}\left(z\right),\label{eq:VMF_distribution}
\end{align}
where $\kappa\geq0$, $\left\Vert \mu\right\Vert =1$, $\mathrm{I}_{a}\left(\cdot\right)$
is the modified Bessel function of the first kind and order $a$,
$\Gamma\left(\cdot\right)$ is the gamma function and $\chi_{A}\left(\cdot\right)$
is the indicator function on set $A$.

\subsection{From pixel to DOA\label{subsec:Pixel-DOA-equivalence}}

In this section, we explain how to convert a pixel $\left(i_{x},i_{y}\right)$
in the image to a DOA $z$ in the camera reference frame. The known
camera parameters are the FoV $\left[f_{x},f_{y}\right]$ in radians,
width $w$ in pixels and height $h$ in pixels. As $w$ and $h$ are
even numbers, see Table \ref{tab:Video-specifications}, the center
pixel of the image is
\begin{align}
\left(c_{x},c_{y}\right) & =\left(\frac{w}{2},\frac{h}{2}\right).
\end{align}

We will test two methods to convert a pixel $\left(i_{x},i_{y}\right)$
into azimuth $\varphi\in\left[0,2\pi\right)$ and elevation $\theta\in\left[-\pi/2,\pi/2\right]$
in the camera reference frame. The first approach is to estimate $\left(\varphi,\theta\right)$
as \cite{Helgesen20,Helgesen22}
\begin{equation}
\varphi=\frac{i_{x}-c_{x}}{w}f_{x},\quad\theta=\frac{i_{y}-c_{y}}{h}f_{y}.\label{eq:phi_theta_pixel_old}
\end{equation}
The second approach makes use of the pinhole camera model \cite{Hartley_book03}
to obtain
\begin{equation}
\varphi=\arctan\frac{i_{x}-c_{x}}{f_{I}},\quad\theta=\arctan\frac{i_{y}-c_{y}}{f_{I}},\label{eq:phi_theta_pixel}
\end{equation}
where $f_{I}$ is the distance (in pixels) from the camera center
to the image plane, which can be estimated as
\begin{align}
f_{I} & =\frac{1}{2}\left(\frac{w}{2\tan\left(f_{x}/2\right)}+\frac{h}{2\tan\left(f_{y}/2\right)}\right).\label{eq:f_I}
\end{align}
The proof of (\ref{eq:phi_theta_pixel}) and (\ref{eq:f_I}), and
the relation with (\ref{eq:phi_theta_pixel_old}), is provided in
Appendix \ref{sec:AppendixA}. 

Using the spherical coordinates, the DOA unit vector corresponding
to $\left(\varphi,\theta\right)$ is $z=\left[z_{x},z_{y},z_{z}\right]^{T}$
with \cite{Edwards_book73}
\begin{align}
z_{x} & =\cos\varphi\cos\theta,\label{eq:z_x}\\
z_{y} & =\sin\varphi\cos\theta,\\
z_{z} & =\sin\theta.\label{eq:z_z}
\end{align}

\subsection{Single-object measurement density\label{subsec:Object-generated-measurement}}

The single-object measurement density $l\left(\cdot|x\right)$ given
an object state $x$ is modelled as VMF (\ref{eq:VMF_distribution})
with
\begin{align}
l\left(z|x\right) & =\mathcal{V}\left(z;h(x),\kappa\right),\label{eq:single_object_VMF_likelihood}
\end{align}
where $h(x)$ is the function that gives the mean direction for an
object with state $x$. Function $h(x)$ depends on the object position
$p=\left[p_{x},p_{y},0\right]^{T}\in\mathbb{R}^{3}$ in the local
coordinate system, the drone location $s\in\mathbb{R}^{3}$ in the
local coordinate system, and the camera frame quaternion $q=q_{1}+q_{2}i+q_{3}j+q_{4}k$,
which indicates the orientation of the camera in the local coordinate
system. For simplicity, we keep the dependence of $h(x)$ on $s$
and $q$ in the notation implicit.

Assuming that the drone is a point object, we obtain  
\begin{align}
h(x) & =\frac{R_{q}\left(p-s\right)}{\left\Vert R_{q}\left(p-s\right)\right\Vert },\label{eq:h_x}
\end{align}
where $R_{q}$ is the rotation matrix of $q$ \cite[pp. 178]{Kuipers_book99}
\begin{align}
R_{q} & =\left(\begin{array}{ccc}
2q_{1}^{2}-1+2q_{2}^{2} & 2q_{2}q_{3}+2q_{1}q_{4} & 2q_{2}q_{4}-2q_{1}q_{3}\\
2q_{2}q_{3}-2q_{1}q_{4} & 2q_{1}^{2}-1+2q_{3}^{2} & 2q_{3}q_{4}+2q_{1}q_{2}\\
2q_{2}q_{4}+2q_{1}q_{3} & 2q_{3}q_{4}-2q_{1}q_{2} & 2q_{1}^{2}-1+2q_{4}^{2}
\end{array}\right).\label{eq:Rotation_matrix}
\end{align}
The numerator in (\ref{eq:h_x}) corresponds to the object position
in the camera coordinate system, so $h(x)$ is the projection of this
vector on the unit sphere.

\subsection{Clutter intensity\label{subsec:Clutter-generated-measurement}}

We model the clutter intensity as uniformly distributed in the camera
FoV. In addition, the average number of clutter measurements is $\overline{\lambda}^{C}$.
As the VMF density (\ref{eq:VMF_distribution}) is defined w.r.t.
the uniform distribution, the clutter intensity must be defined w.r.t.
the uniform distribution on the sphere. 

Defining $\varphi\left(z\right)$ and $\theta\left(z\right)$ to be
the azimuth and elevation of a measurement $z$, see (\ref{eq:z_x})-(\ref{eq:z_z}),
the camera FoV is $\mathbb{F}=\left\{ z:\left\Vert z\right\Vert =1,\varphi\left(z\right)\in f_{x}/2\cdot\left[-1,1\right],\theta\left(z\right)\in f_{y}/2\cdot\left[-1,1\right]\right\} $.
Then, the clutter intensity is
\begin{align}
\lambda^{C}\left(z\right) & =\frac{\overline{\lambda}^{C}}{u^{C}}\chi_{z\in\mathbb{F}}\left(z\right),\label{eq:clutter_intensity}\\
u^{C} & =\frac{f_{x}\sin\left(f_{y}/2\right)}{2\pi},\label{eq:normalising_constant_clutter}
\end{align}
where $u^{C}$ is the normalising constant of the uniform distribution
on the FoV. The proof of (\ref{eq:normalising_constant_clutter})
is provided in Appendix \ref{sec:AppendixB}. It should be noted that
if the FoV were the whole sphere ($f_{x}=2\pi$, $f_{y}=\pi),$ then
$u^{C}=1$, as the density is defined w.r.t. the uniform distribution. 

\subsection{Multi-object measurement model\label{subsec:Multi-object-measurement-model}}

Given an image, we use a neural network to obtain a set of bounding
boxes with the object detections \cite{Luo21}. We refer to this neural
network as the object detection network. We then obtain the DOAs of
the center of the bounding boxes to form the set $\mathbf{z}_{k}$
of measurements. That is, given the bounding box parameterisation
$\left[b_{x},b_{y},b_{w},b_{h}\right]^{T}$, where $\left[b_{x},b_{y}\right]^{T}$
is its upper left corner, $b_{w}$ its width, and $b_{h}$ its height,
the center $\left[b_{c,x},b_{c,y}\right]^{T}$ is
\begin{equation}
b_{c,x}=b_{x}+b_{w}/2,\quad b_{c,y}=b_{y}+b_{h}/2,
\end{equation}
which can be converted to a DOA, see Section \ref{subsec:Pixel-DOA-equivalence}.

We then assume that these object-generated DOA measurements follow
the VMF distribution (\ref{eq:single_object_VMF_likelihood}), clutter
has intensity (\ref{eq:clutter_intensity}) and the probability of
detection is a constant. With this measurement model, we can now perform
the TPMBM update step from the drone images. How to select the parameters
of this measurement model is explained in Section \ref{sec:Measurement-parameter-estimation}.

\section{Measurement model parameter estimation\label{sec:Measurement-parameter-estimation}}

Once we have trained the neural network detector, we estimate the
measurement model parameters $p^{D}$, $\kappa$, and $\overline{\lambda}^{C}$.
To do so, we propose a coordinate ascent method \cite{Nocedal_book99}
to maximise the likelihood after we introduce auxiliary variables
that represent the data associations. We present the form of the likelihood
in Section \ref{subsec:Likelihood}. The optimisations over the auxiliary
variables and measurement model parameters are provided in Sections
\ref{subsec:Optimising_a} and \ref{subsec:Optimising_measurement_model},
respectively.

\subsection{Likelihood\label{subsec:Likelihood}}

For each frame $k=1,...,K$, we know the sets of DOAs $\mathbf{y}_{k}=\left\{ h\left(x_{k}^{1}\right),...,h\left(x_{k}^{n_{k}}\right)\right\} $,
where $\mathbf{x}_{k}=\left\{ x_{k}^{1},...,x_{k}^{n_{k}}\right\} $
is the set of ground truth objects. Set $\mathbf{y}_{k}$ is obtained
by mapping the center pixel of each manually annotated bounding box
to a DOA, see Section \ref{subsec:Pixel-DOA-equivalence}. We also
know the set of measurements $\mathbf{z}_{k}=\left\{ z_{k}^{1},...,z_{k}^{m_{k}}\right\} $
obtained by a neural network. 

By applying the convolution formula \cite{Mahler_book14}, the likelihood
of $\mathbf{z}_{k}$ given $\mathbf{x}_{k}$ is Poisson multi-Bernoulli
\cite{Williams15b}. That is, $\mathbf{z}_{k}$ is the union of independent
PPP clutter and $n_{k}$ possible object-generated measurements. The
density is
\begin{align}
p\left(\mathbf{z}_{k}|\mathbf{x}_{k}\right) & =e^{-\overline{\lambda}^{C}}\sum_{\mathbf{z}_{k}^{0}\uplus...\uplus\mathbf{z}_{k}^{n_{k}}=\mathbf{z}_{k}}\left[\prod_{z\in\mathbf{z}_{k}^{0}}\lambda^{C}\left(z\right)\right]\prod_{i=1}^{n_{k}}t\left(\mathbf{z}_{k}^{i}|x_{k}^{i}\right),\label{eq:measurement_likelihood}
\end{align}
where the Bernoulli density of the object-generated measurements is
\begin{align}
t\left(\mathbf{z}_{k}^{i}|x_{k}^{i}\right) & =\begin{cases}
p^{D}l(z|x_{k}^{i}) & \mathbf{z}_{k}^{i}=\{z\}\\
1-p^{D} & \mathbf{z}_{k}^{i}=\emptyset\\
0 & \mathrm{otherwise}.
\end{cases}\label{eq:Bernoulli_generated_likelihood}
\end{align}
In (\ref{eq:measurement_likelihood}), the sum goes through all disjoint,
and possibly empty, sets $\mathbf{z}_{k}^{0}$,...,$\mathbf{z}_{k}^{n_{k}}$
such that their union is $\mathbf{z}_{k}$. 

To make parameter optimisation easier, we remove the convolution sum
in (\ref{eq:measurement_likelihood}) by the introduction of auxiliary
variables. An auxiliary variable $a_{k}^{j}\in\mathbb{\mathbb{A}}_{k}=\left\{ 0,1,...,n_{k}\right\} $
for each measurement $z_{k}^{j}$ is defined such that an extended
measurement is $\widetilde{z}_{k}^{j}=\left(a_{k}^{j},z_{k}^{j}\right)\in\mathbb{\mathbb{A}}_{k}\times S^{2}$,
and $a_{k}^{j}=0$ if it is a clutter measurement and $a_{k}^{j}=i$
if it is a measurement generated by $x_{k}^{i}$ \cite{Angel20_e,Kim22}. 

Given a set of measurements with auxiliary variables $\widetilde{\mathbf{z}}_{k}$,
we denote its subsets with each auxiliary variable by $\widetilde{\mathbf{z}}_{k}^{0},...,\widetilde{\mathbf{z}}_{k}^{n_{k}}$.
Then, the likelihood with auxiliary variables becomes
\begin{align}
\widetilde{p}\left(\widetilde{\mathbf{z}}_{k}|\mathbf{x}_{k}\right) & =e^{-\overline{\lambda}^{C}}\left[\prod_{\left(0,z\right)\in\widetilde{\mathbf{z}}_{k}^{0}}\lambda^{C}\left(z\right)\right]\prod_{i=1}^{n_{k}}\widetilde{t}^{i}\left(\widetilde{\mathbf{z}}_{k}^{i}|x_{k}^{i}\right),\label{eq:measurement_likelihood-auxiliary}\\
\widetilde{t}^{i}\left(\widetilde{\mathbf{z}}_{k}^{i}|x_{k}^{i}\right) & =\begin{cases}
p^{D}l(z|x_{k}^{i}) & \widetilde{\mathbf{z}}_{k}^{i}=\{\left(i,z\right)\}\\
1-p^{D} & \widetilde{\mathbf{z}}_{k}^{i}=\emptyset\\
0 & \mathrm{otherwise}.
\end{cases}\label{eq:Bernoulli_generated_likelihood_aux}
\end{align}
Integrating out the auxiliary variables in (\ref{eq:measurement_likelihood-auxiliary}),
we obtain (\ref{eq:measurement_likelihood}). It should be noted that
the likelihood with auxiliary variables (\ref{eq:measurement_likelihood-auxiliary})
is a lower bound of the likelihood (\ref{eq:measurement_likelihood}),
as all terms in (\ref{eq:measurement_likelihood}) are positive and
(\ref{eq:measurement_likelihood-auxiliary}) only contains one of
them. It should be noted that to evaluate (\ref{eq:measurement_likelihood_all}),
it is sufficient to know the set $\mathbf{y}_{k}$ of DOAs instead
of $\mathbf{x}_{k}$.

Assuming that all measurements are within the FoV, the likelihood
for all time steps is
\begin{align}
\widetilde{p}\left(\widetilde{\mathbf{z}}_{1:K}|\mathbf{x}_{1:K}\right) & =e^{-K\overline{\lambda}^{C}}\prod_{k=1}^{K}\left[\left(\frac{\overline{\lambda}^{C}}{u^{C}}\right)^{|\widetilde{\mathbf{z}}_{k}^{0}|}\prod_{i=1}^{n_{k}}t^{i}\left(\widetilde{\mathbf{z}}_{k}^{i}|x_{k}^{i}\right)\right]\label{eq:measurement_likelihood_all}
\end{align}
where $\widetilde{\mathbf{z}}_{1:K}=\left(\widetilde{\mathbf{z}}_{1},...,\widetilde{\mathbf{z}}_{K}\right)$
and $\mathbf{x}_{1:K}=\left(\mathbf{x}_{1},...,\mathbf{x}_{K}\right)$. 

We proceed to iteratively optimise (\ref{eq:measurement_likelihood_all})
over the variables $a_{k}=\left(a_{k}^{1},...,a_{k}^{m_{k}}\right)$,
$\overline{\lambda}^{C}$, $p_{D}$ and $\kappa$. 

\subsection{Optimising $a_{k}$\label{subsec:Optimising_a}}

Fixing $\overline{\lambda}^{C}$, $p_{D}$ and $\kappa$, we seek
the values of $a_{k}$ for $k=1,...,K$ that maximise (\ref{eq:measurement_likelihood_all}).
The solution can be found by solving an independent optimal assignment
problem at each time step. 

For time step $k$, we create an $m_{k}\times\left(n_{k}+m_{k}\right)$
cost matrix $C_{k}$ with value
\begin{align}
 & -\ln\left(\begin{array}{cccccc}
\frac{p^{D}l(z_{k}^{1}|x_{k}^{1})}{1-p^{D}} & ... & \frac{p^{D}l(z_{k}^{1}|x_{k}^{n_{k}})}{1-p^{D}} & \frac{\overline{\lambda}^{C}}{u^{C}} & 0 & 0\\
\vdots & \vdots & \vdots & 0 & \vdots & 0\\
\frac{p^{D}l(z_{k}^{m_{k}}|x_{k}^{1})}{1-p^{D}} & ... & \frac{p^{D}l(z_{k}^{m_{k}}|x_{k}^{n_{k}})}{1-p^{D}} & 0 & 0 & \frac{\overline{\lambda}^{C}}{u^{C}}
\end{array}\right).
\end{align}

The optimal assignment can be written as an $m_{k}\times\left(n_{k}+m_{k}\right)$
binary matrix $S_{k}$ whose rows sum to one and its columns to either
zero or one. Then, we can obtain the optimal assignment by minimising
$\mathrm{tr}\left(S_{k}^{T}C_{k}\right)$ using the Hungarian algorithm
\cite{Blackman_book99} or alternative methods \cite{Crouse16}. The
optimal value of $S_{k}$ can then be written as an auxiliary variable
$a_{k}$. 

\subsection{Optimising $\overline{\lambda}^{C}$, $p^{D}$ and $\kappa$\label{subsec:Optimising_measurement_model}}

The optimisation over $\overline{\lambda}^{C}$, $p^{D}$ and $\kappa$
is provided in the following lemma.
\begin{lem}
\label{lem:Optimisation}Given the auxiliary variables $a_{k}$, $k=1,...,K$,
the values of $\overline{\lambda}^{C}$, $p^{D}$ and $\kappa$ that
maximise the likelihood (\ref{eq:measurement_likelihood_all}) are
\begin{align}
\overline{\lambda}^{C} & =\frac{\sum_{k=1}^{K}|\mathbf{\widetilde{z}}_{k}^{0}|}{K},\label{eq:lambda_clutter_opt}\\
p^{D} & =\frac{\sum_{k=1}^{K}\sum_{i=1:\widetilde{\mathbf{z}}_{k}^{i}=\{z\}}^{n_{k}}1}{\sum_{k=1}^{K}n_{k}},\label{eq:p_d_opt}\\
\frac{\mathrm{I}_{3/2}\left(\kappa\right)}{\mathrm{I}_{1/2}\left(\kappa\right)} & =\overline{r},\label{eq:kappa_opt_exact}
\end{align}
where
\begin{align}
\overline{r} & =\frac{\sum_{k=1}^{K}\left[\sum_{i=1:\widetilde{\mathbf{z}}_{k}^{i}=\{z\}}^{n_{k}}\left(h\left(x_{k}^{i}\right)^{T}z\right)\right]}{\sum_{k=1}^{K}\sum_{i=1:\widetilde{\mathbf{z}}_{k}^{i}=\{z\}}^{n_{k}}1}.
\end{align}

\end{lem}
Lemma \ref{lem:Optimisation} is proved in Appendix \ref{sec:AppendixC}.
The optimal clutter rate (\ref{eq:lambda_clutter_opt}) is the average
number of clutter measurements per time step, as indicated by the
auxiliary variables. Equation (\ref{eq:p_d_opt}) is the empirical
probability of detection for the assignments provided by the auxiliary
variables. That is, the numerator is the total number of detected
objects across all time steps and the denominator is the total number
of objects across all time steps. Finally, $\kappa$ is obtained by
solving (\ref{eq:kappa_opt_exact}), where $\overline{r}$ is a measure
of the similarity of the associated DOA measurements w.r.t. $h\left(x_{k}^{i}\right)$.

For $\overline{r}\geq0.9$, which is the case for a detector with
sufficiently good performance, an accurate approximate solution to
(\ref{eq:kappa_opt_exact}) is \cite[Eq. (10.3.7)]{Mardia_book00}
\begin{align}
\kappa & \simeq\frac{1}{1-\overline{r}}.\label{eq:eq:kappa_opt_approx}
\end{align}

\section{Multi-object dynamic model\label{sec:Multi-object-dynamic-model}}

This section explains the multi-object dynamic model, which consists
of the probability of survival $p^{S}$, the single-object transition
density $g\left(\cdot\left|x\right.\right)$, and birth intensity
$\lambda_{k}^{B}\left(\cdot\right)$. The models used are based on
prior knowledge on the system set-up.

\subsection{Transition density}

To model vehicle dynamics, we adopt a nearly constant velocity model
\cite{Sarkka_book13}. The single-object transition density is 
\begin{align}
g\left(\cdot\left|x\right.\right) & =\mathcal{N}\left(\cdot;Fx,Q\right),\label{eq:transition_density}
\end{align}
\begin{align}
F=I_{2}\otimes\left(\begin{array}{cc}
1 & \tau\\
0 & 1
\end{array}\right),\;Q=\sigma_{q}^{2}I_{2}\otimes\left(\begin{array}{cc}
\tau^{3}/3 & \tau^{2}/2\\
\tau^{2}/2 & \tau
\end{array}\right),
\end{align}
where $\tau$ is the sampling time between video frames, $F$ is the
transition matrix and $Q$ the process noise covariance matrix with
parameter $\sigma_{q}^{2}$, which indicates the intensity of the
changes in the velocity. The probability of survival is set as a constant.

\subsection{Birth intensity }

The birth intensity at time step $k$ is Gaussian
\begin{align}
\lambda_{k}^{B}\left(x\right) & =\overline{\lambda}_{k}^{B}\mathcal{N}\left(x;\overline{x}_{k}^{B},P_{k}^{B}\right),
\end{align}
where $\overline{\lambda}_{k}^{B}$ is the expected number of new
born objects, $\overline{x}_{k}^{B}$ is the mean and $P_{k}^{B}$
the covariance matrix. We proceed to explain how we calculate $\overline{x}_{k}^{B}$
and $P_{k}^{B}$.

We consider that objects may appear anywhere in the FoV with a uniform
angular distribution in the interval $\left[-f_{x}/2,f_{x}/2\right]\times\left[-f_{y}/2,f_{y}/2\right]$.
This distribution has zero-mean and covariance 
\begin{align}
R^{B} & =\frac{1}{12}\left(\begin{array}{cc}
f_{x}^{2} & 0\\
0 & f_{y}^{2}
\end{array}\right).
\end{align}

We then map the mean and covariance of this distribution into the
object positional space. To do so, we select $m$ sigma points $\mathcal{Y}_{1},...,\mathcal{Y}_{m}$
and weights $\omega_{1},...,\omega_{m}$ with zero-mean and covariance
matrix using a sigma-point method \cite{Sarkka_book13}, such as the
unscented transform \cite{Julier04}. These points are transformed
to DOAs $\mathcal{Z}_{1},...,\mathcal{Z}_{m}$ using (\ref{eq:z_x})-(\ref{eq:z_z}),
and afterwards to positional points $\mathcal{P}_{1},...,\mathcal{P}_{m}$
using the equations in Appendix \ref{sec:AppendixD}. The resulting
positional mean and covariance are \cite{Sarkka_book13}
\begin{align}
\overline{p}_{k}^{B} & \approx\sum_{j=1}^{m}\omega_{j}\mathcal{P}_{j},\\
P_{p,k}^{B} & \approx\sum_{j=1}^{m}\omega_{j}\left(\mathcal{P}_{j}-\overline{p}_{k}^{B}\right)\left(\mathcal{P}_{j}-\overline{p}_{k}^{B}\right)^{T}.
\end{align}
Then, $\overline{x}_{k}^{B}$ has the positional elements in $\overline{p}_{k}^{B}$
and zero velocity. Matrix $P_{k}^{B}$ has the positional elements
in $P_{p,k}^{B}$ and the variance of the velocity elements are set
to $\sigma_{v}^{2}$. 

\section{Gaussian TPMBM implementation\label{sec:Gaussian-TPMBM-implementation}}

In this work, following \cite{Angel20_e,Angel22_c}, we use a Gaussian
implementation of the TPMBM filter for the set of all trajectories.
In this setting, the single-trajectory densities $p_{k'|k}^{i,a^{i}}\left(\cdot\right)$
in (\ref{eq:Bernoulli_density_filter}) have a deterministic start
time and are Gaussian for each possible end time, and the PPP intensity
$\lambda_{k'|k}\left(\cdot\right)$ in (\ref{eq:PPP_intensity}) is
a Gaussian mixture that only considers alive trajectories. We proceed
to explain the main aspects of the implementation, and refer the reader
to \cite{Angel20_e,Angel22_c} for all the details.

\subsection{Single-trajectory density}

A Gaussian density of a trajectory starting at time step $\tau$ with
mean $\overline{x}$ and covariance $P$ is denoted by \cite{Angel20_e}
\begin{align}
\mathcal{N}\left(t,x^{1:\nu};\tau,\overline{x},P\right) & =\begin{cases}
\mathcal{N}\left(x^{1:\nu};\overline{x},P\right) & t=\tau,\,\nu=\iota\\
0 & \mathrm{otherwise},
\end{cases}\label{eq:Trajectory_Gaussian}
\end{align}
where the trajectory length is $\iota=\mathrm{dim}\left(\overline{x}\right)/n_{x}$,
with $\mathrm{dim}\left(\overline{x}\right)$ being the dimensionality
of vector $\overline{x}$. A trajectory with density (\ref{eq:Trajectory_Gaussian})
starts at time step $\tau$ and ends at time step $\tau+\iota-1$.
If $k$ is the current time step, for an alive trajectory, $\tau+\iota-1=k$,
and for a dead trajectory $\tau+\iota-1<k$. 

In each $p_{k'|k}^{i,a^{i}}\left(\cdot\right)$, see (\ref{eq:Bernoulli_density_filter}),
of the Gaussian TPMBM for all trajectories \cite{Angel20_e}, the
birth time is known $t^{i}$, and the trajectory may end at any time
between its birth and the current time step. The single-trajectory
density $p_{k'|k}^{i,a^{i}}\left(\cdot\right)$ is then of the form
\begin{align}
p_{k'|k}^{i,a^{i}}\left(X\right) & =\sum_{l=t^{i}}^{k'}\beta_{k'|k}^{i,a^{i}}\left(l\right)\mathcal{N}\left(X;t^{i},\overline{x}_{k'|k}^{i,a^{i}}\left(l\right),P_{k'|k}^{i,a^{i}}\left(l\right)\right),\label{eq:single_trajectory_Gaussian_all}
\end{align}
where $\beta_{k'|k}^{i,a^{i}}\left(l\right)$ is the probability that
the trajectory ends at time step $l$, and $\overline{x}_{k'|k}^{i,a^{i}}\left(l\right)\in\mathbb{R}^{\iota n_{x}}$
and $P_{k'|k}^{i,a^{i}}\left(l\right)\in\mathbb{R}^{\iota n_{x}\times\iota n_{x}}$,
with $\iota=l-t^{i}+1$, are the mean and covariance given that the
trajectory ends at time step $l$. The weights $\beta_{k'|k}^{i,a^{i}}\left(l\right)$
sum to one over $l$.

\subsection{Single-trajectory prediction and update}

The multi-object dynamic model, see Section \ref{sec:Multi-object-dynamic-model},
is linear and Gaussian, so we can directly apply the linear/Gaussian
single-trajectory prediction \cite{Angel20_e}. The measurement model,
see Section \ref{sec:Camera-as-direction-of-arrival}, is non-linear/non-Gaussian
so we need to perform a Gaussian approximation in the update. Specifically,
in the update, we first perform a Gaussian approximation to the posterior
of the current object state, and then we update past states of the
trajectories using Gaussian properties \cite{Angel22_c,Beutler09}.

To perform the update with detection hypotheses, we use the IPLF applied
to the VMF distribution \cite{Angel19_c}. In addition, we use the
improvement of the normalising constant approximation explained in
\cite{Angel21_b}. 

\subsection{Practical considerations}

Implementing the TPMBM filter also requires the following practical
considerations. To deal with the ever increasing number of global
hypotheses, we make use of ellipsoidal gating \cite{Blackman_book99},
Bernoulli pruning and global hypothesis pruning, in which we discard
Bernoulli components with low probability of existence and global
hypotheses with low weight \cite{Granstrom20}. We also use Murty's
algorithm to select global hypotheses with high weight in the update
\cite{Murty68,Angel18_b}. 

As time goes on, trajectories may become increasingly long and it
becomes computationally intractable to deal with the single-trajectory
density (\ref{eq:single_trajectory_Gaussian_all}). Therefore, we
make use of the $L$-scan approximation \cite{Angel20_e}, in which
trajectory states before the last $L$ time steps are considered independent.
This implies that the largest covariance matrices are of size $L\times L$
and the implementation can deal with long time sequences. 

\section{Experiments\label{sec:Experiments}}

In this section, we analyse the performance of the proposed vehicle
monitoring system with synthetic and experimental data obtained with
a Parrot Anafi USA drone in Liverpool. We compare different implementations
of the PMBM and TPMBM filters via numerical simulations. Then, we
test the accuracy of the DOA camera measurements in Section \ref{subsec:Accuracy-of-DOA-measurements}.
Finally, we evaluate the filters using experimental data from the
drone with optical and thermal cameras in Section \ref{subsec:MOT-performance}. 

\subsection{Synthetic experiments\label{subsec:Synthetic-experiments}}

We compare different implementations of the PMBM and TPMBM filters
in a synthetic experiment. Both filters have been implemented with
the following parameters to perform the single-object updates: L0N1,
L0N5 and L1N5, where L$x$N$y$ indicates that $x=1$ if there is
likelihood improvement, or zero otherwise, and $y$ is the maximum
number of IPLF iterations \cite{Angel21_b}. The IPLF uses the unscented
transform with weight 1/3 at the origin \cite{Julier04} and the Kullback-Leibler
divergence stopping criterion with threshold $10^{-2}$ \cite[Eq. (30)]{Angel15_c}.
It should be noted that the implementations with L0N1, no likelihood
improvement, and 1 IPLF iteration, correspond to the UKF implementations
\cite{Morelande06}, providing a baseline to assess the improvement
provided by the IPLF. In turn, UKF implementations generally outperform
the extended Kalman filter implementations so these are not considered
\cite{Julier04,Morelande13}.

The filters prune Bernoulli components with existence lower than $10^{-4}$,
prune MBM components with weight lower than $10^{-4}$, and prune
PPP components with weight lower than $10^{-5}$. The ellipsoidal
gating threshold is 50 and the maximum number of global hypothesis
is 100, obtained via Murty's algorithm \cite{Murty68,Angel18_b}.
In addition, the TPMBM filter has been implemented with $L=1$ and
$L=5$, and parameter $\Gamma_{a}=10^{-3}$ \cite{Angel20_e}.

We consider a scenario, similar to \cite{Williams15b}, with 101 time
steps in which 4 objects get in close proximity and then separate
and one object disappears at time step 51. Objects move with $p^{S}=0.99$,
$\sigma_{q}^{2}=0.5$ ($\mathrm{m^{2}/s^{3}}$) and $\tau=1/6\,\mathrm{s}$,
and are born with $\overline{\lambda}_{0}^{B}=1$ and $\overline{\lambda}_{k}^{B}=0.025$
with $\sigma_{v}^{2}=20^{2}$ ($\mathrm{m}^{2}/\mathrm{s}^{2}$).
The drone is located at the origin of the local coordinate system
with an elevation of 25 m. The camera is pointing at point $\left(25,25,0\right)$
(m) in the local coordinate system. The VMF measurements have $\kappa=700$
and are generated using the code in \cite{Chen15_b}. The clutter
intensity parameter is $\overline{\lambda}^{C}=5$. 

\begin{figure}
\begin{centering}
\includegraphics[scale=0.3]{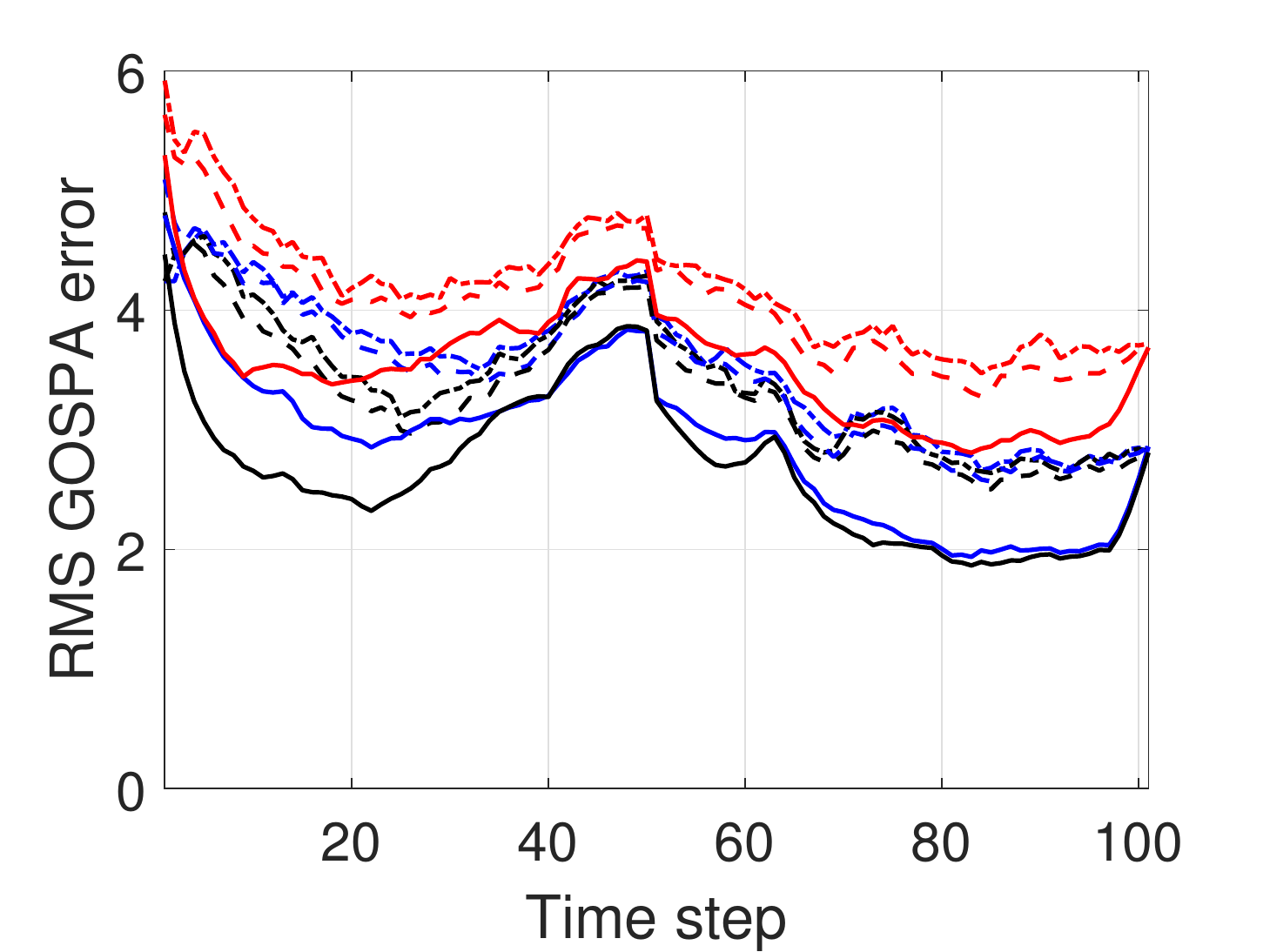}\includegraphics[scale=0.3]{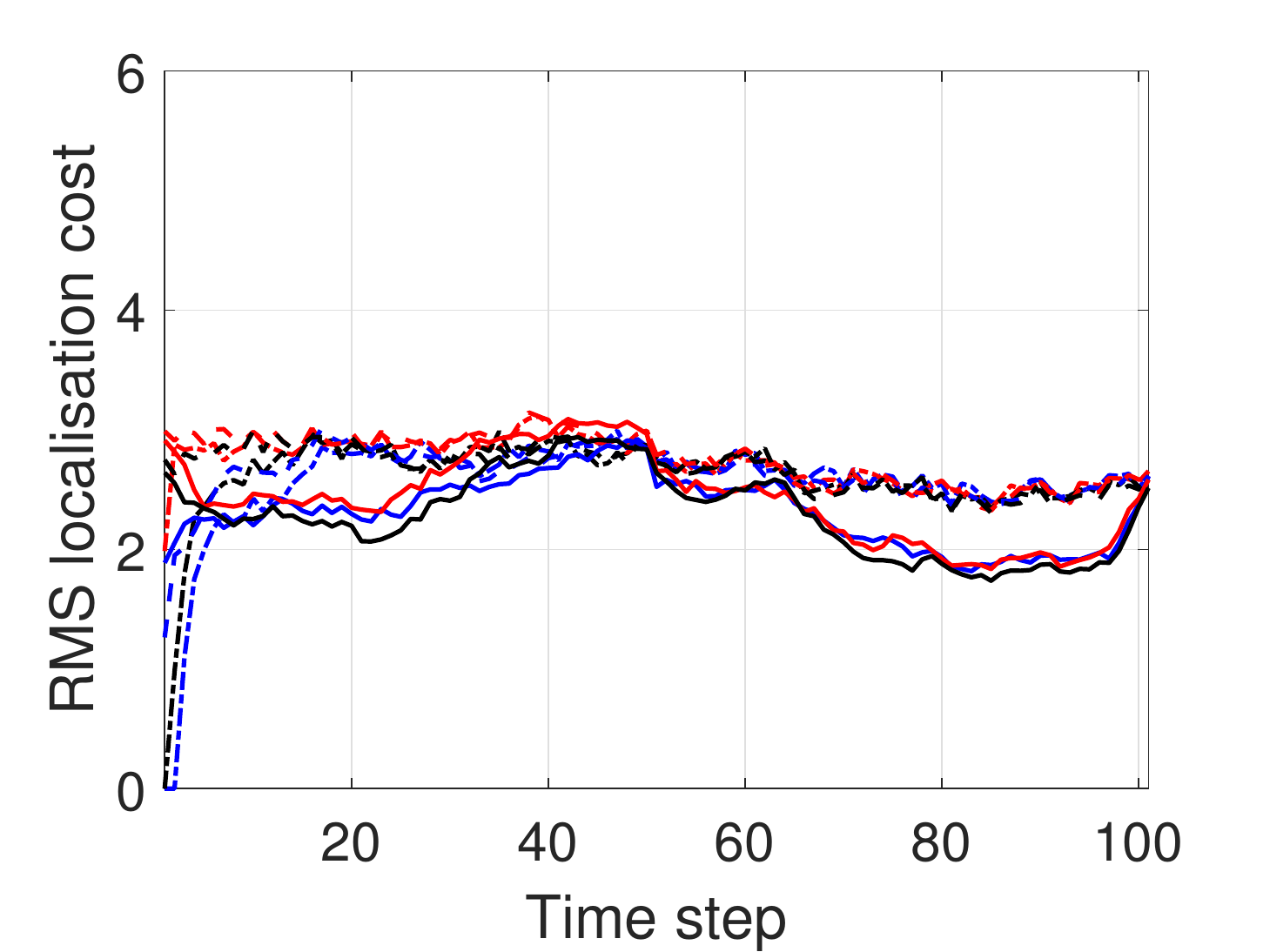}
\par\end{centering}
\begin{centering}
\includegraphics[scale=0.3]{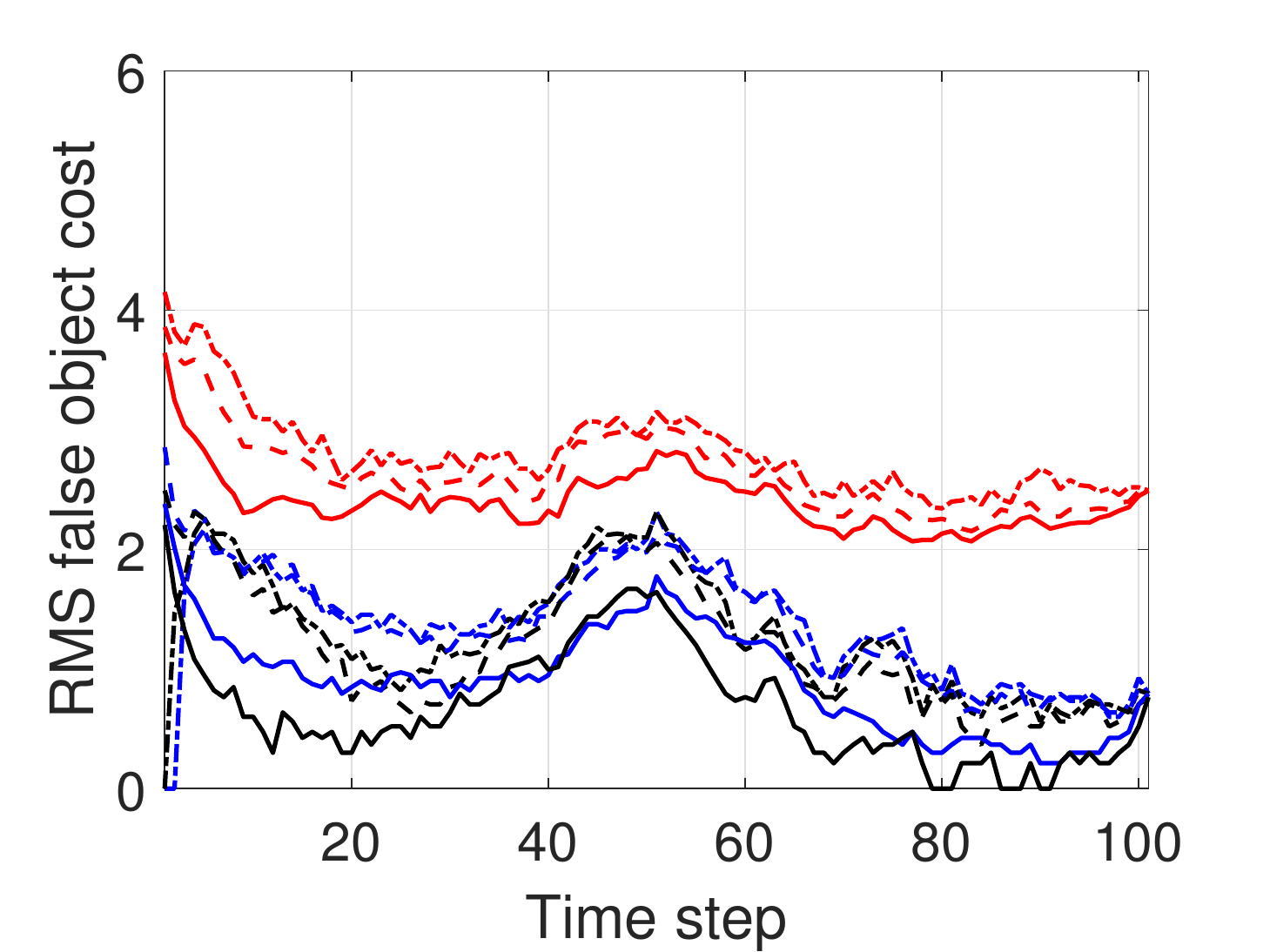}\includegraphics[scale=0.3]{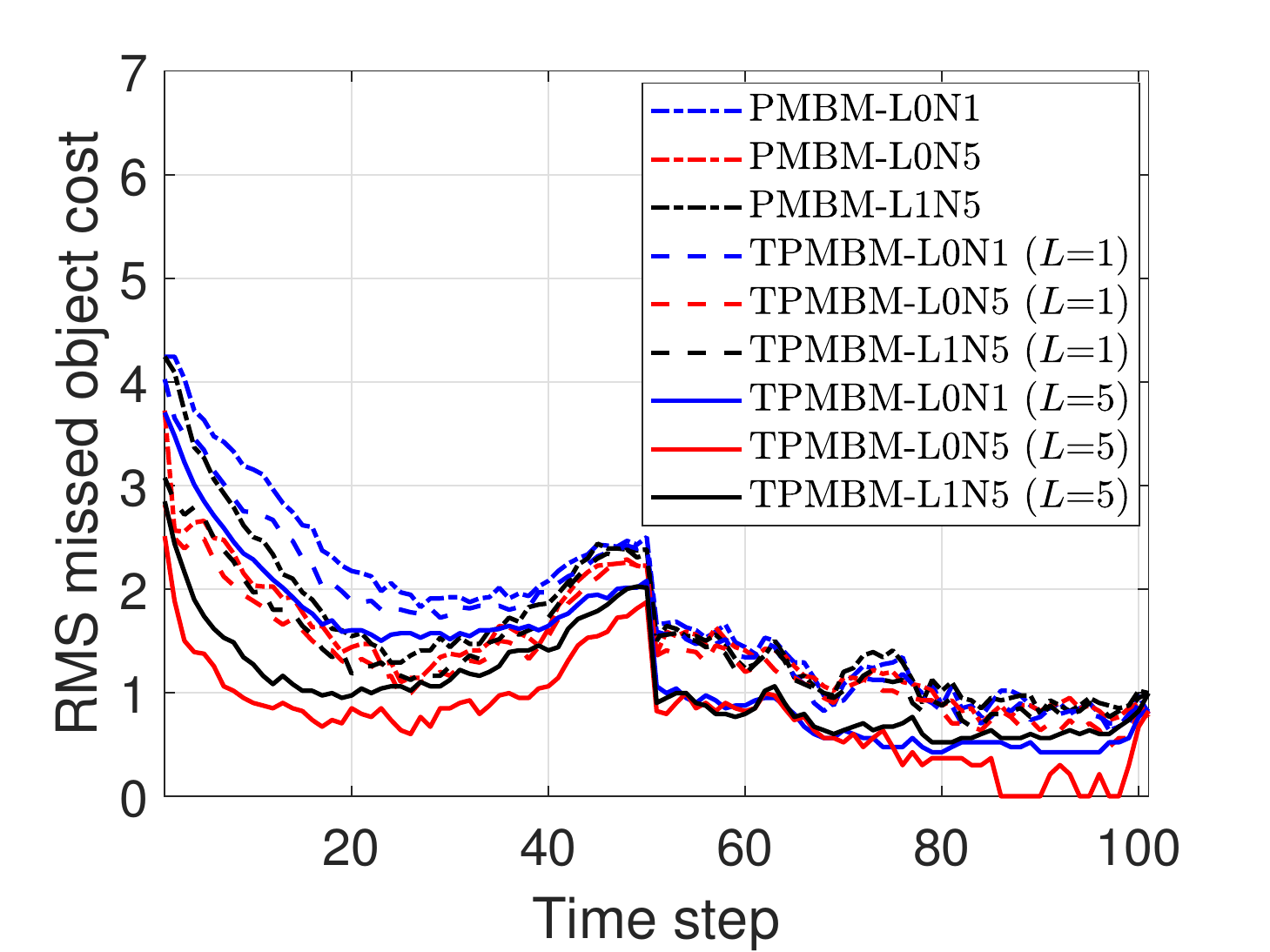}
\par\end{centering}
\caption{\label{fig:RMS-GOSPA-error-synthetic}RMS-GOSPA error (m) and its
decomposition. The TPMBM filter with $L=5$, likelihood improvement,
and 5 IPLF iterations is the best performing filter.}
\end{figure}

We evaluate filter performance via Monte Carlo simulation with 100
runs. We use the generalised optimal sub-pattern assignment (GOSPA)
metric with parameters $\alpha=2$, $p=2$ and $c=3\,\mathrm{m}$
to estimate the error of the ground truth set of vehicle positions
and the estimated positions at each time step at the end of the simulation
\cite{Rahmathullah17}. The root mean square GOSPA (RMS GOSPA) at
each time step is shown in Figure \ref{fig:RMS-GOSPA-error-synthetic}.
As expected, the best performing filter is TPMBM-L1N5 with $L=5$,
as it uses likelihood improvement, 5 IPLF iterations and an $L$-scan
window of 5. Lowering the $L$-scan window decreases performance as
it does not improve estimation of past states.  The PMBM has worse
performance than TPMBM as it estimates object states sequentially
without modifying past estimates. For the PMBM using 5 IPLF iterations
and likelihood improvement also lowers the error compared to the UKF
update (L0N1). 

The computational times to run one Monte Carlo simulation of the algorithms
on an Intel core i5 laptop are provided in Table \ref{tab:Computational-times-synthetic}.
The PMBM filter with L0N1 is the fastest filter. A higher number of
IPLF iterations and $L$ increase the computational burden.

\begin{table}

\caption{\label{tab:Computational-times-synthetic}Computational times in seconds
of the algorithms }

\begin{centering}
\begin{tabular}{c|c|c|c}
\hline 
 &
\multicolumn{1}{c|}{PMBM} &
\multicolumn{1}{c|}{TPMBM ($L=1$)} &
\multicolumn{1}{c}{TPMBM ($L=5$)}\tabularnewline
\hline 
L0N1 &
3.4 &
4.0 &
4.6\tabularnewline
L0N5 &
5.9 &
9.4 &
10.1\tabularnewline
L1N5 &
6.6 &
7.5 &
7.9\tabularnewline
\hline 
\end{tabular}
\par\end{centering}
\end{table}

\subsection{Accuracy of DOA camera measurements\label{subsec:Accuracy-of-DOA-measurements}}

In this section, we evaluate the accuracy of the estimated distances
on the ground using the drone optical camera as a DOA sensor. Given
a pixel on the image, we can obtain its DOA, see Section \ref{sec:Camera-as-direction-of-arrival},
and then we can obtain the corresponding position on the ground, see
Appendix \ref{sec:AppendixD}. Then, we can estimate distances on
the ground and evaluate how accurate these estimates are. The accuracy
of an estimated distance on the ground depends on several factors
such as the accuracy in the drone elevation, camera quaternion, lens
distortion, and gimbal stability, which depends on the drone motion
and external factors such as wind speed \cite{Liu20}. 

\begin{figure}
\begin{centering}
\includegraphics[scale=0.36]{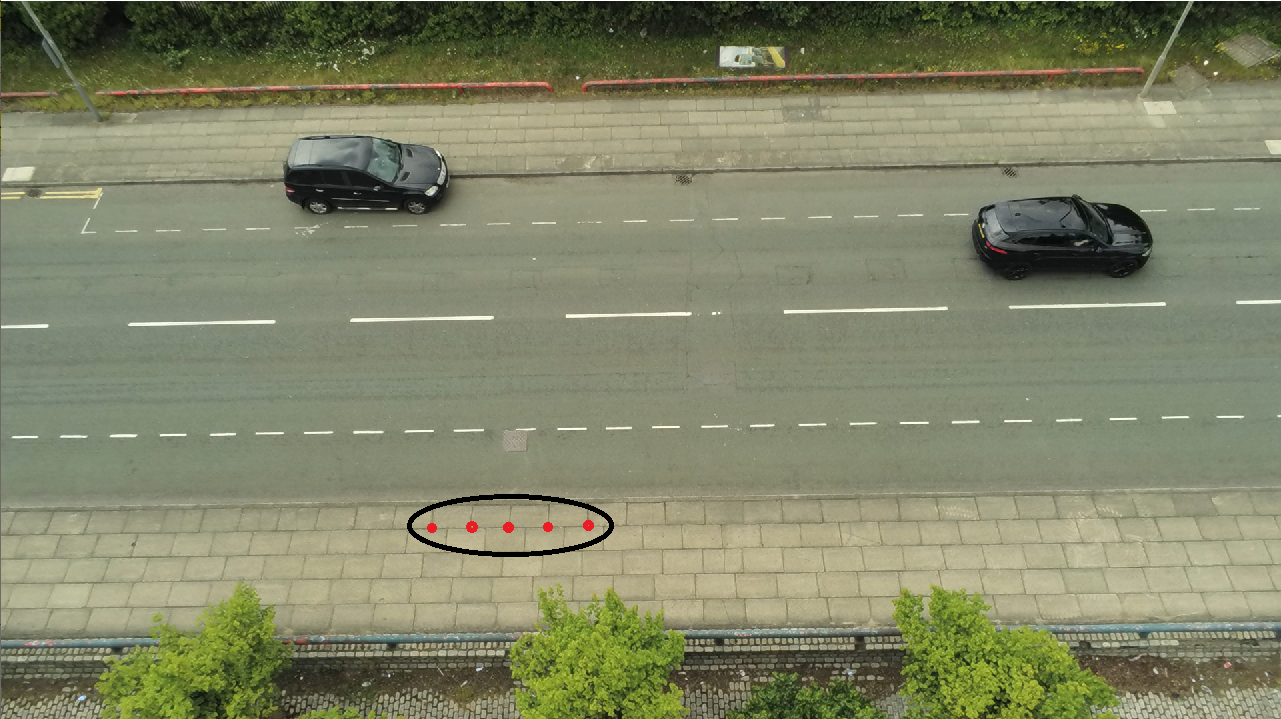}
\par\end{centering}
\caption{\label{fig:Landmarks_drone}Image from the drone (elevation 20.8 m).
The 5 points/landmarks on the pavement marked with red dots (inside
the black ellipse) are used to measure the accuracy of the drone sensing
system. }
\end{figure}

The scenario is shown in Figure \ref{fig:Landmarks_drone}. The drone
takes off and reaches an elevation of around 30 meters, while tilting
the camera to point at the road. The length of each pavement tile
measured with a tape measure is 0.9 m. We then take 5 consecutive
points defined by these tiles to act as landmarks, as indicated in
Figure \ref{fig:Landmarks_drone}. 

We identify the pixels representing the landmarks in 8 different frames
of the video, with 200 frames in between these frames (which correspond
to 6.7 seconds, see Table \ref{tab:Video-specifications}). We have
chosen these frames as they are evenly distributed while the drone
is flying up and the camera is being tilted. We then calculate the
distance of the corresponding points in the local coordinate system.
We calculate the root mean square error (RMSE) of the estimated distances
between consecutive landmarks, which are located 0.9 meters apart.
We use two methods to calculate the DOA from a pixel: Method 1 uses
(\ref{eq:phi_theta_pixel_old}) and Method 2 uses (\ref{eq:phi_theta_pixel}).

The resulting RMSEs are shown in Table \ref{tab:RMSE-landmarks2},
where we also show the angles of arrival (azimuth $\varphi$ and elevation
$\theta$) for the first landmark, obtained via Method 2. We can see
that the RMSE stays below 0.08 m for both methods for a range of elevations
and DOA. The RMSE remains unchanged from the 8-th frame on Table \ref{tab:RMSE-landmarks2}
until the end of the video (5.8 seconds after the 8-th frame). Across
all frames, Method 2 generally performs better than Method 1 and is
therefore the method we use in the next section.

\begin{table}
\caption{\label{tab:RMSE-landmarks2}RMSE of the distances between consecutive
landmarks}

\centering{}%
\begin{tabular}{c|>{\centering}p{1cm}cc>{\centering}p{1.6cm}>{\centering}p{1.6cm}}
\hline 
Frame &
Elevation (m) &
$\varphi\,({}^{\circ})$ &
$\theta\,({}^{\circ})$ &
RMSE Method 1 (m) &
RMSE Method 2 (m)\tabularnewline
\hline 
1 &
11.0 &
-7.8 &
3.7 &
0.07 &
0.05\tabularnewline
2 &
17.1 &
-6.5 &
7.6 &
0.03 &
0.08\tabularnewline
3 &
20.6 &
-5.5 &
9.3 &
0.06 &
0.06\tabularnewline
4 &
20.6 &
-10.4 &
10.2 &
0.08 &
0.03\tabularnewline
5 &
23.7 &
-9.5 &
8.2 &
0.07 &
0.03\tabularnewline
6 &
26.8 &
-10.2 &
6.5 &
0.07 &
0.03\tabularnewline
7 &
29.4 &
-9.4 &
9.6 &
0.08 &
0.03\tabularnewline
8 &
31.5 &
-9.3 &
8.4 &
0.07 &
0.04\tabularnewline
\hline 
\end{tabular}
\end{table}

\subsection{Experimental MOT performance\label{subsec:MOT-performance}}

In this section, we analyse the performance of the proposed PMBM and
TPMBM filters using recorded videos with the drone. The PMBM and TPMBM
filters have also been implemented with the linear and Gaussian (LG)
measurement likelihood used in \cite{Helgesen20}, which uses (\ref{eq:phi_theta_pixel_old}).
With the LG measurement likelihood, we perform a linear and Gaussian
update by projecting the measurement on the ground plane. The measurement
noise covariance matrix is obtained using the unscented transform
with a standard deviation for the pixels of 30 \cite{Helgesen23}.
We have also implemented a probability hypothesis density (PHD) filter
and a cardinality PHD (CPHD) filter on the image plane \cite{Maggio_book11,Mahler_book14}
based on the neural network detections. The estimated outputs of these
filters are then projected to the ground plane, providing two additional
baseline algorithms for comparison. The PHD/CPHD filters use a 4-D
state vector with the position and velocity on the image plane. For
the PHD/CPHD filters, the measurement noise covariance matrix is $9I_{2}$,
where $I_{n}$ is the identity matrix of size $n$, and the transition
density is (\ref{eq:transition_density}) with $\sigma_{q}^{2}=9\,(\mathrm{pixel^{2}/s^{3}})$.
The PHD/CPHD birth model is Gaussian with the mean at the center of
the image and covariance matrix $P_{p}^{B}=\mathrm{diag}\left(\left[w^{2}/4,1,h^{2}/4,1\right]\right)$.

For each frame in the video, we have manually annotated the bounding
box for each object. The available labelled data does not contain
the sequence of bounding boxes of each object across all frames, which
would be required to train a deep learning MOT algorithm \cite{Xu20,Milan16_arxiv}.
The considered multi-object dynamic model is the same as in Section
\ref{subsec:Synthetic-experiments} but with $\tau=1/30\,\mathrm{s}$
for the optical camera and $\tau=1/8.6\,\mathrm{s}$ for the thermal
camera, see Table \ref{tab:Video-specifications}.

The ground truth set of object positions at each time step is considered
to be the mapping of the center of each annotated bounding box in
each frame to the ground plane, using (\ref{eq:phi_theta_pixel})
and Appendix \ref{sec:AppendixD}. Then, after we have processed all
frames with the filters, we estimate the set of trajectories and evaluate
positional errors. To do so, we compute the GOSPA metric $d_{G}\left(\cdot,\cdot\right)$
with parameters $\alpha=2$, $p=2$ and $c=3$ m at each time step,
and then compute the following RMS GOSPA error across all time steps
\begin{align}
d\left(\mathbf{x}_{1:K},\hat{\mathbf{x}}_{1:K}\right) & =\sqrt{\frac{1}{K}\sum_{k=1}^{K}d_{G}^{2}\left(\mathbf{x}_{k},\hat{\mathbf{x}}_{k}\right)},\label{eq:error_GOSPA_time}
\end{align}
where $\mathbf{x}_{k}$ is the ground truth set of targets at time
step $k$, and $\hat{\mathbf{x}}_{k}$ is its estimate. It should
be noted that (\ref{eq:error_GOSPA_time}) corresponds to the metric
for sets of trajectories in \cite{Angel20_d}, with the track switching
cost tending to zero, and normalised by the time window. The metric
in \cite{Angel20_d} could only be computed explicitly if we had manually
annotated the ground truth set of trajectories. Nevertheless, in traffic
monitoring, objects tend to follow the road in an orderly fashion
so a low number of track switches are expected in most situations.
Therefore, in these scenarios, (\ref{eq:error_GOSPA_time}) provides
an accurate approximation of the trajectory metric \cite{Angel20_d},
even though there is no ground truth trajectory data. For example,
in the synthetic experiments in Section \ref{tab:Computational-times-synthetic},
the difference between the RMS trajectory metric error (with switching
cost $\gamma=1$) is roughly 0.02 higher than (\ref{eq:error_GOSPA_time})
for all filters, without affecting the ranking of the algorithms. 

In the rest of the subsection, we first explain the object detection
network and then present the results with the optical and thermal
cameras. 

\subsubsection{Object detection network\label{subsec:Object-detection-network}}

Any object detection network can be used in combination with our algorithm.
In this experimental validation, we use a YOLOv4-tiny detector \cite{Wang21}
due to its speed for real-time object detection and accuracy. The
network only considers one object class ``Vehicle''. We have manually
annotated the bounding boxes of the considered videos using the Matlab
Video Labeler app, helping the process with the integrated KLT point
tracker algorithm \cite{Lucas81}. We have used 6 optical camera videos
and 6 thermal videos, all obtained in daylight conditions. The lengths
of the videos and the ranges of drone elevations are provided in Table
\ref{tab:Length-elevation-videos}.
\begin{center}
\begin{table}
\caption{\label{tab:Length-elevation-videos}Length and range of drone elevations
of the videos}

\centering{}%
\begin{tabular}{c|cc|cc}
\hline 
 &
\multicolumn{2}{c|}{Optical} &
\multicolumn{2}{c}{Thermal}\tabularnewline
\hline 
Video &
Length (s) &
Elevation (m)  &
Length (s)  &
Elevation (m)\tabularnewline
\hline 
1 &
23 &
34 - 34 &
12 &
29 - 30\tabularnewline
2 &
109 &
14 - 35 &
30 &
30 - 31\tabularnewline
3 &
98 &
28 - 40 &
43 &
41 - 45\tabularnewline
4 &
96 &
35 - 35 &
39 &
35 - 36\tabularnewline
5 &
50 &
40 - 40 &
69 &
40 - 40\tabularnewline
6 &
32 &
36 - 36 &
45 &
31 - 31\tabularnewline
\hline 
\end{tabular}
\end{table}
\par\end{center}

The training of the detector for optical images has been performed
by re-training a pre-trained model on the COCO data set. Re-training
has been performed with 3 videos. We have used an input size of $416\times416\times3$
and 6 anchor boxes, whose sizes are estimated from the training data
\cite{Redmon17}. We have used data augmentation with random horizontal
flipping, color jittering, X/Y scaling, and rotations from $-90^{\circ}$
to $90^{\circ}$. The optimisation algorithm is the Adam optimiser,
with initial learning rate 0.001, mini-batch size of 8, moving batch
normalisation statistics, and a maximum number of epochs of 200. For
the thermal images, we have re-trained the obtained detector for optical
images using 3 thermal videos. We have performed this re-training
as the features of the vehicles on both type of images are quite similar. 

In addition, for each video frame, we have also extracted the latitude,
longitude, elevation and camera quaternion of the drone from the video
metadata using ExifTool\footnote{https://exiftool.org}. These are
required to establish the coordinate systems, see Section \ref{subsec:Coordinate-systems},
and run the filters. 

\subsubsection{Optical camera}

We use video 4 to perform the measurement model parameter estimation,
see Section \ref{sec:Measurement-parameter-estimation}. The resulting
parameters are $p^{D}=0.95$, $\kappa=65000$, and $\overline{\lambda}^{C}=0.05$.
The RMS GOSPA errors (\ref{eq:error_GOSPA_time}) for the algorithms
are provided in Table \ref{tab:RMS-GOSPA-errors-optical}. We can
see that the TPMBM ($L=5$) with 5 IPLF iterations and likelihood
improvement is the best performing filter in general. This result
is to be expected as it is the filter with highest performance in
theory, and can improve the estimation of past states of the trajectories.
The PMBM filter implementations with L1N5 generally have worse performance
than the TPMBM implementations. Overall, the proposed PMBM and TPMBM
filter implementations outperform the LG implementations and the PHD/CPHD
filters on the image plane.

\begin{table*}
\caption{\label{tab:RMS-GOSPA-errors-optical}RMS GOSPA errors across time
in the optical videos }

\centering{}%
\begin{tabular}{c|c|c|cccc|cccc|cccc}
\hline 
 &
PHD &
CPHD &
\multicolumn{4}{c|}{PMBM} &
\multicolumn{4}{c|}{TPMBM ($L=1$)} &
\multicolumn{4}{c}{TPMBM ($L=5$)}\tabularnewline
\hline 
Video &
- &
- &
LG &
L0N1 &
L0N5 &
L1N5 &
LG &
L0N1 &
L0N5 &
L1N5 &
LG &
L0N1 &
L0N5 &
L1N5\tabularnewline
\hline 
1 &
1.38 &
1.44 &
1.70 &
1.92 &
0.82 &
0.87 &
1.66 &
2.18 &
0.79 &
0.79 &
1.63 &
2.02 &
\uline{0.77} &
0.78\tabularnewline
2 &
1.46 &
1.47 &
2.31 &
1.06 &
0.92 &
0.89 &
2.32 &
1.19 &
0.90 &
0.89 &
2.27 &
1.12 &
\uline{0.86} &
\uline{0.86}\tabularnewline
3 &
1.44 &
1.28 &
4.48 &
0.70 &
0.73 &
0.73 &
4.48 &
0.76 &
0.70 &
0.70 &
4.47 &
\uline{0.69} &
\uline{0.69} &
\uline{0.69}\tabularnewline
4 &
1.38 &
1.39 &
2.23 &
1.10 &
0.86 &
0.83 &
2.23 &
1.21 &
0.85 &
0.85 &
2.19 &
1.16 &
0.83 &
\uline{0.82}\tabularnewline
5 &
2.40 &
2.46 &
3.01 &
1.92 &
1.97 &
1.92 &
3.01 &
1.97 &
1.90 &
1.88 &
2.98 &
1.90 &
1.88 &
\uline{1.86}\tabularnewline
6 &
3.17 &
3.20 &
4.16 &
2.88 &
2.87 &
2.82 &
4.16 &
2.90 &
2.83 &
2.82 &
4.11 &
2.80 &
\uline{2.79} &
\uline{2.79}\tabularnewline
\hline 
\end{tabular}
\end{table*}

\subsubsection{Thermal camera}

\begin{figure}
\begin{centering}
\includegraphics[scale=0.3]{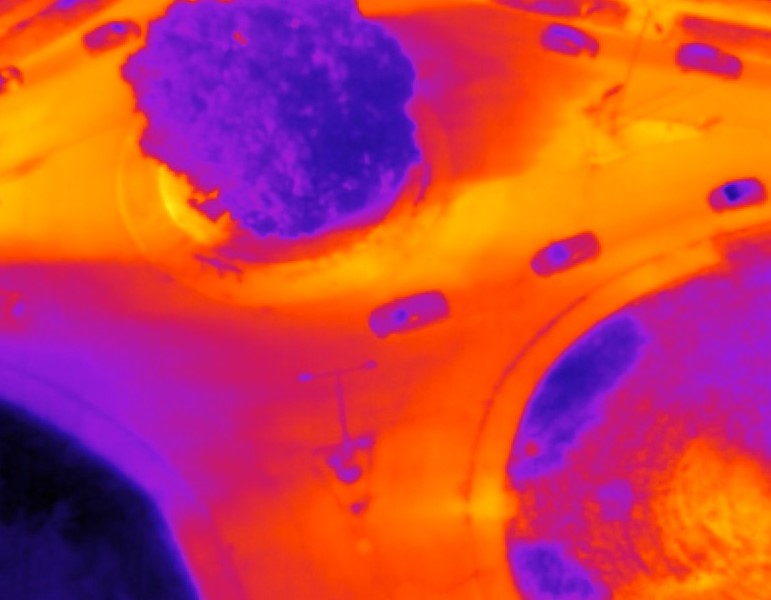}
\par\end{centering}
\caption{\label{fig:Example-thermal_image}Example of a thermal image of a
roundabout with 7 vehicles, one of them entering the image (elevation
44.7 m). }
\end{figure}

An example of a thermal image from the drone is shown in Figure \ref{fig:Example-thermal_image}.
Using thermal video 4, the measurement model parameter estimation
algorithm provides $p^{D}=0.99$ $\kappa=58800$, and $\overline{\lambda}^{C}=0.07$.
The RMS GOSPA errors (\ref{eq:error_GOSPA_time}) for the algorithms
are provided in Table \ref{tab:RMS-GOSPA-errors-thermal}. As with
the optical camera, the TPMBM filter is the best performing filter.
The TPMBM implementation with $L=5$, likelihood improvement and 5
IPLF iterations generally provides the best estimates of the vehicle
trajectories. The TPMBM with $L=1$, no likelihood improvement and
1 IPLF iteration performs slightly worse. The PMBM filter shows a
decrease in performance compared to the TPMBM filter. The LG implementations
and the PHD/CPHD filters on the image plane do not perform as well
as the proposed implementations of the TPMBM and PMBM filters.

\begin{table*}
\caption{\label{tab:RMS-GOSPA-errors-thermal}RMS GOSPA errors across time
in the thermal videos}

\centering{}%
\begin{tabular}{c|c|c|cccc|cccc|cccc}
\hline 
 &
PHD &
CPHD &
\multicolumn{4}{c|}{PMBM} &
\multicolumn{4}{c|}{TPMBM ($L=1$)} &
\multicolumn{4}{c}{TPMBM ($L=5$)}\tabularnewline
\hline 
Video &
- &
- &
LG &
L0N1 &
L0N5 &
L1N5 &
LG &
L0N1 &
L0N5 &
L1N5 &
LG &
L0N1 &
L0N5 &
L1N5\tabularnewline
\hline 
1 &
1.53 &
1.80 &
1.30 &
0.76 &
0.82 &
0.72 &
1.29 &
0.74 &
\uline{0.66} &
0.72 &
1.34 &
0.70 &
\uline{0.66} &
0.72\tabularnewline
2 &
1.72 &
1.97 &
1.39 &
0.82 &
1.05 &
0.81 &
1.36 &
0.77 &
0.73 &
0.66 &
1.35 &
0.73 &
0.72 &
\uline{0.65}\tabularnewline
3 &
2.34 &
2.53 &
2.51 &
1.57 &
1.68 &
1.56 &
2.50 &
1.54 &
1.52 &
1.51 &
2.47 &
\uline{1.49} &
1.50 &
\uline{1.49}\tabularnewline
4 &
2.31 &
2.51 &
2.06 &
1.75 &
1.78 &
1.74 &
2.04 &
1.71 &
1.68 &
1.68 &
2.06 &
1.68 &
\uline{1.67} &
\uline{1.67}\tabularnewline
5 &
1.45 &
1.43 &
0.99 &
0.75 &
0.93 &
0.76 &
0.96 &
0.63 &
\uline{0.61} &
\uline{0.61} &
1.04 &
0.62 &
0.62 &
0.62\tabularnewline
6 &
3.23 &
3.24 &
2.88 &
2.82 &
2.96 &
2.82 &
2.86 &
2.80 &
2.80 &
2.79 &
2.89 &
2.80 &
2.79 &
\uline{2.78}\tabularnewline
\hline 
\end{tabular}
\end{table*}

\section{Conclusions\label{sec:Conclusions}}

This paper has proposed a multi-object tracking algorithm for vehicle
monitoring using a drone equipped with a camera. The camera is modelled
as a DOA sensor with its uncertainty modelled as a VMF distribution.
We have derived the corresponding measurement model, which requires
mapping the vehicle position on the ground into a direction of arrival,
making use of the drone camera pose. 

The multi-object tracking algorithm is based on the PMBM filter for
sets of trajectories, which provides us with the optimal Bayesian
solution for trajectory estimation under the standard multi-object
tracking models. To apply this detection-based filter, vehicles in
the image are detected using a neural network, retrained with labelled
images obtained by the drone. 

We have also derived a measurement model parameter estimation algorithm
that enables us to obtain the probability of detection, clutter intensity
and concentration parameter that best fit a lower bound on the likelihood.
This algorithm can also be used to obtain the best parameters for
different conditions, such as changing illumination or weather.

The algorithm has been first tested via numerical simulations. We
have also carried out an experimental validation to test the accuracy
of estimated distances on the ground, and the vehicle tracking accuracy
using optical and thermal cameras.

\appendices{}

\section{\label{sec:AppendixA}}

In this appendix, we prove (\ref{eq:phi_theta_pixel}) and (\ref{eq:f_I})
using the pinhole camera model \cite{Hartley_book03} assuming that
pixels are evenly distributed on the image plane. We also show the
connection with (\ref{eq:phi_theta_pixel_old}). From the camera specifications,
see Table \ref{tab:Video-specifications}, we know the FoV along the
horizontal and vertical axes ($f_{x}$ and $f_{y}$ in radians) as
well as the width $w$ and height $h$ in pixels. Using the pinhole
camera model and just considering $f_{x}$ and $h$ (horizontal axis),
we obtain the estimate $f_{I,x}$ of $f_{I}$ as \cite{Hartley_book03}
\begin{align}
f_{I,x} & =\frac{w}{2\tan\left(f_{x}/2\right)}.
\end{align}
Performing the same calculation for the vertical axis, we obtain $f_{I,y}$
by substituting $h$ and $f_{y}$ instead of $w$ and $f_{x}$ in
the previous equation. Equation (\ref{eq:f_I}) is the average between
$f_{I,x}$ and $f_{I,y}$. 

Using the geometry of the pinhole camera model, we obtain
\begin{align}
\tan\varphi= & \frac{i_{x}-c_{x}}{f_{I}}.\label{eq:tan_phi}
\end{align}
The same process can be used to obtain $\theta$ and prove (\ref{eq:phi_theta_pixel}). 

Considering the values in Table \ref{tab:Video-specifications}, we
obtain $f_{I,x}=1396.81$ and $f_{I,y}=1396.90$ for the optical camera,
which are very similar values. If we use $f_{I,x}$ instead of $f_{I}$
in for $\varphi$ and $f_{I,y}$ instead of $f_{I}$ in for $\theta$
in (\ref{eq:phi_theta_pixel}), and we approximate $\tan x\approx x$,
valid for sufficiently small $x$, we obtain (\ref{eq:phi_theta_pixel_old}).

\section{\label{sec:AppendixB}}

In this appendix, we calculate the normalising constant $u^{C}$ of
the clutter intensity, see (\ref{eq:normalising_constant_clutter}).
To do so, we integrate the differential spherical area element (surface
element) on the unit sphere \cite{Edwards_book73} in the region of
the camera FoV. In addition, as the VMF density is defined w.r.t.
the uniform distribution, see (\ref{eq:VMF_distribution}), we need
to divide the previous integral by $4\pi$, which is the area of the
unit sphere. That is, we have that 
\begin{align}
u^{C} & =\frac{1}{4\pi}\int_{-f_{x}/2}^{f_{x}/2}\int_{\pi/2-f_{y}/2}^{\pi/2+f_{y}/2}\sin\theta d\varphi d\theta.
\end{align}
This integral can be solved analytically yielding (\ref{eq:normalising_constant_clutter}).

\section{\label{sec:AppendixC}}

In this appendix, we prove Lemma \ref{lem:Optimisation}. The log
likelihood can be obtained from (\ref{eq:measurement_likelihood_all}),
(\ref{eq:Bernoulli_generated_likelihood_aux}) and (\ref{eq:VMF_distribution})
as
\begin{align}
L & =\ln\widetilde{p}\left(\widetilde{\mathbf{z}}_{1:K}|\mathbf{x}_{1:K}\right)\nonumber \\
 & =\sum_{k=1}^{K}\left[|\widetilde{\mathbf{z}}_{k}^{0}|\ln\left(\frac{\overline{\lambda}^{C}}{u^{C}}\right)+\sum_{i=1:|\widetilde{\mathbf{z}}_{k}^{i}|=\emptyset}^{n_{k}}\ln\left(1-p^{D}\right)\right.\nonumber \\
 & +\sum_{i=1:\widetilde{\mathbf{z}}_{k}^{i}=\{z\}}^{n_{k}}\left(\ln p^{D}+1/2\ln\frac{\kappa}{2}+\kappa h\left(x_{k}^{i}\right)^{T}z\right.\nonumber \\
 & \left.\left.-\ln\mathrm{I}_{1/2}\left(\kappa\right)-\ln\Gamma\left(3/2\right)\right)\vphantom{\ln\left(\frac{\overline{\lambda}^{C}}{u^{C}}\right)}\right]-K\overline{\lambda}^{C}.\label{eq:log_likelihood_app}
\end{align}
We proceed to maximise (\ref{eq:log_likelihood_app}) w.r.t. $\overline{\lambda}^{C}$,
$p^{D}$ and $\kappa$ in the following subsections.

\subsection{Optimisation of (\ref{eq:log_likelihood_app}) w.r.t. $\overline{\lambda}^{C}$}

The derivative of (\ref{eq:log_likelihood_app}) w.r.t. $\overline{\lambda}^{C}$
is
\begin{align}
\frac{\partial L}{\partial\overline{\lambda}^{C}} & =-K+\frac{1}{\overline{\lambda}^{C}}\sum_{k=1}^{K}|\mathbf{\widetilde{z}}_{k}^{0}|.\label{eq:Derivative_lambda_c}
\end{align}
Equating (\ref{eq:Derivative_lambda_c}) to zero, we obtain (\ref{eq:lambda_clutter_opt}).
Taking the second derivative, we can check (\ref{eq:lambda_clutter_opt})
is a maximum.

\subsection{Optimisation of (\ref{eq:log_likelihood_app}) w.r.t. $p^{D}$}

The derivative of (\ref{eq:log_likelihood_app}) w.r.t. $p^{D}$ is
\begin{align}
\frac{\partial L}{\partial p_{D}} & =\frac{-1}{1-p^{D}}\left[\sum_{k=1}^{K}\sum_{i=1:|\widetilde{\mathbf{z}}_{k}^{i}|=\emptyset}^{n_{k}}1\right]\nonumber \\
 & +\frac{1}{p^{D}}\left[\sum_{k=1}^{K}\sum_{i=1:\widetilde{\mathbf{z}}_{k}^{i}=\{z\}}^{n_{k}}1\right].\label{eq:Derivative_pd}
\end{align}
Equating (\ref{eq:Derivative_pd}) to zero, we obtain (\ref{eq:p_d_opt}).
Taking the second derivative, we can check (\ref{eq:p_d_opt}) is
a maximum.

\subsection{Optimisation of (\ref{eq:log_likelihood_app}) w.r.t. $\kappa$}

The derivative of (\ref{eq:log_likelihood_app}) w.r.t. $\kappa$
is
\begin{align}
\frac{\partial L}{\partial\kappa} & =\sum_{k=1}^{K}\left[\sum_{i=1:\widetilde{\mathbf{z}}_{k}^{i}=\{z\}}^{n_{k}}\left(h\left(x_{k}^{i}\right)^{T}z\right)\right]\nonumber \\
 & +\left(\frac{1}{2\kappa}-\frac{\frac{\partial\mathrm{I}_{1/2}\left(\kappa\right)}{\partial\kappa}}{\mathrm{I}_{1/2}\left(\kappa\right)}\right)\sum_{k=1}^{K}\sum_{i=1:\widetilde{\mathbf{z}}_{k}^{i}=\{z\}}^{n_{k}}1.\label{eq:Derivative_kappa}
\end{align}
Equating (\ref{eq:Derivative_kappa}) to zero and applying the equality
\cite[Eq. (A.8)]{Mardia_book00}
\begin{align}
\frac{\partial\mathrm{I}_{1/2}\left(\kappa\right)}{\partial\kappa} & =\frac{1}{2\kappa}\mathrm{I}_{1/2}\left(\kappa\right)+\mathrm{I}_{3/2}\left(\kappa\right),
\end{align}
we obtain that the stationary point for $\kappa$ meets (\ref{eq:kappa_opt_exact}).
Then, considering that $\mathrm{I}_{3/2}\left(\kappa\right)/\mathrm{I}_{1/2}\left(\kappa\right)$
is a strictly increasing function \cite{Schou78}, the second derivative
is negative, which proves that the stationary point is a maximum.

\section{\label{sec:AppendixD}}

In this appendix, given a pixel $\left(i_{x},i_{y}\right)$, we calculate
the corresponding position $\left[p_{x},p_{y},0\right]^{T}$ of an
object on the ground in the local coordinate system. We first obtain
the DOA $z$ in the camera reference frame using (\ref{eq:phi_theta_pixel})
and (\ref{eq:z_x})-(\ref{eq:z_z}). The DOA vector in the local coordinate
system $\nu=\left[\nu_{x},\nu_{y},\nu_{z}\right]^{T}$ is calculated
with the inverse rotation of (\ref{eq:Rotation_matrix}) such that
\begin{align}
\nu & =\left(R_{q}\right)^{-1}z.
\end{align}
Then, given the sensor position $\left[s_{x},s_{y},s_{z}\right]^{T}$
in the local coordinate system, the point on the ground $\left[p_{x},p_{y},0\right]^{T}$
with this DOA meets
\begin{equation}
\left[s_{x},s_{y},s_{z}\right]^{T}+\lambda\left[\nu_{x},\nu_{y},\nu_{z}\right]^{T}=\left[p_{x},p_{y},0\right]^{T},\label{eq:append_intersection}
\end{equation}
where $\lambda\in\mathbb{R}$. In (\ref{eq:append_intersection}),
the left-hand side represents the line defined by the DOA.

Solving for $\lambda$ in (\ref{eq:append_intersection}) yields that
the position on the ground has $(p_{x},p_{y})$ coordinates
\begin{equation}
p_{x}=s_{x}-\frac{s_{z}}{\nu_{z}}\nu_{x},\quad p_{y}=s_{y}-\frac{s_{z}}{\nu_{z}}\nu_{y}.
\end{equation}

\selectlanguage{british}%
\bibliographystyle{IEEEtran}
\bibliography{7C__Trabajo_laptop_Referencias_Referencias}
\selectlanguage{english}%

\end{document}